\documentclass{article} 
\usepackage{iclr2023_conference,times}

\usepackage{amsmath,amsfonts,bm}

\newcommand{\eqdef}{\stackrel{\text{def}}{=}}

\newcommand{\specialcell}[2][c]{%
  \begin{tabular}[#1]{@{}c@{}}  #2\end{tabular}}

\newcommand{\todo}[1]{\textcolor{red}{ [TODO] #1}}

\def\eqref#1{equation~\ref{#1}}

\def\1{\bm{1}}

\DeclareMathAlphabet{\mathsfit}{\encodingdefault}{\sfdefault}{m}{sl}
\SetMathAlphabet{\mathsfit}{bold}{\encodingdefault}{\sfdefault}{bx}{n}

\DeclareMathOperator*{\argmax}{arg\,max}

\usepackage{url}
\usepackage{wrapfig}
\usepackage{float}
\usepackage{amsthm}
\usepackage{mathtools, etoolbox}
\usepackage{array}
\usepackage{comment}
\usepackage{paralist} 
\usepackage{bm}
\usepackage{algorithm}
\usepackage{algpseudocode}
\usepackage{graphicx}
\usepackage{caption}
\usepackage{subcaption}
\definecolor{customcolor}{HTML}{004496}
\usepackage[colorlinks,allcolors=customcolor]{hyperref}

\usepackage{environ}
\NewEnviron{myequation}{%
\begin{equation*}
\scalebox{1.2}{$\BODY$}
\end{equation*}
}

\title{Interpretability in the Wild: a Circuit for\\Indirect Object Identification in GPT-2 small} 

\author{\\ \textbf{Kevin Wang}$^1$\textbf{,} \textbf{Alexandre Variengien}$^1$\textbf{,} \textbf{Arthur Conmy}$^1$\textbf{,} \textbf{Buck Shlegeris}$^1$ \textbf{\&} \textbf{Jacob Steinhardt}$^{1, 2}$ \\
$^1$Redwood Research\\
$^2$UC Berkeley\\
\texttt{kevin@rdwrs.com, alexandre@rdwrs.com,} \\ 
\texttt{arthur@rdwrs.com, buck@rdwrs.com, jsteinhardt@berkeley.edu}
}

\renewcommand{\S}{\textcolor{red}{S }}

\newcommand\ac[1]{\textcolor{blue}{[AC: #1]}}

\newcommand{\pioi}{\mathsf{p}_{\mathrm{IOI}}}

\newcommand{\pabc}{\mathsf{p}_{\mathrm{ABC}}}
\newcommand{\xnew}{x_{\mathrm{new}}}
\newcommand{\xorig}{x_{\mathrm{orig}}}

\definecolor{s_color}{RGB}{201,165,247}
\definecolor{io_color}{RGB}{114,255,100}
\definecolor{end_color}{RGB}{250,210,0}
\iclrfinalcopy 
\makeatother
\renewcommand{\headrulewidth}{0pt}
\begin{document}
\pagenumbering{arabic}

\maketitle

\begin{abstract}
Research in mechanistic interpretability seeks to explain behaviors of machine learning (ML) models in terms of their internal components. However, most previous work either focuses on simple behaviors in small models or describes complicated behaviors in larger models with broad strokes. In this work, we bridge this gap by presenting an explanation for how GPT-2 small performs a natural language task called indirect object identification (IOI). Our explanation encompasses 26 attention heads grouped into 7 main classes, which we discovered using a combination of interpretability approaches relying on causal interventions.
To our knowledge, this investigation is the largest end-to-end attempt at reverse-engineering a natural behavior ``in the wild" in a language model.  We evaluate the reliability of our explanation using three quantitative criteria--\emph{faithfulness, completeness} and \emph{minimality}. Though these criteria support our explanation, they also point to remaining gaps in our understanding. 
Our work provides evidence that a mechanistic understanding of large ML models is feasible, pointing toward opportunities to scale our understanding to both larger models and more complex tasks. Code for all experiments is available at \url{https://github.com/redwoodresearch/Easy-Transformer}.

\end{abstract}


\section{Introduction}
\label{sec:intro}



 Transformer-based language models \citep{Attention,GPT3} have demonstrated an impressive suite of capabilities but largely remain black boxes. Understanding these models is difficult because they employ complex non-linear interactions in densely-connected layers and operate in a high-dimensional space. Despite this, they are already deployed in high-impact settings \citep{zhang2022shifting, caldarini2022literature},  underscoring the urgency of understanding and anticipating possible model behaviors. Some researchers have argued that interpretability is critical for the safe deployment of advanced machine learning systems \citep{Hendrycks2022XRiskAF}.

Work in mechanistic interpretability aims to discover, understand, and verify the algorithms that model weights implement by reverse engineering model computation into human-understandable components \citep{AnthropicMechanisticEssay, meng2022locating, geiger2021causal, geva2020transformer}. By understanding underlying mechanisms, we can better predict out-of-distribution behavior \citep{mu2020compositionalExplanations}, identify and fix model errors \citep{hernandez2021natural,vig2020investigating}, and understand emergent behavior \citep{meca_interp_grokking,barak2022hidden,Wei2022EmergentAO}.

In this work, we aim to mechanistically understand how GPT-2 small \citep{GPT2} implements a simple natural language task. We do so by using \emph{circuit analysis} \citep{interp_survey}, identifying an induced subgraph of the model's computational graph that is human-understandable and responsible for completing the task.




To discover the circuit, we introduce a systematic approach that iteratively traces important components back from the logits, using a causal intervention that we call ``path patching". We supplement this mainline approach with projections in the embedding space, attention pattern analysis, and activation patching to understand the behavior of each component.

We focus on understanding a specific natural language task  
that we call indirect object identification (IOI). In IOI, sentences such as ``When Mary and John went to the store, John gave a drink to'' should be completed with ``Mary''. 
 We chose this task because it is linguistically meaningful and admits an interpretable algorithm: of the two names in the sentence, predict the name that isn't the subject of the last clause. 

We discover a circuit of 26 attention heads--1.1\% of the total number of (head, token position) pairs--that completes the bulk of this task. 
The circuit uses 7 categories of heads (see Figure~\ref{fig:the_circuit}) to implement the algorithm. Together, these heads route information between different name tokens, to the end position, and finally to the output. 
Our work provides, to the best of our knowledge, the most detailed attempt at reverse-engineering an end-to-end behavior in a transformer-based language model. 

By zooming in on a crisp task in a particular model, we obtained several insights about the challenges of mechanistic interpretability. In particular:

\setdefaultleftmargin{1em}{2.2em}{1.87em}{1.7em}{1em}{1em}
\begin{compactitem}
    \item We identified several instances of heads implementing redundant behavior \citep{michel2019sixteen}. The most surprising were ``Backup Name-Mover Heads'', which copy names to the correct position in the output, but only when regular Name-Mover Heads are ablated. This complicates the search for complete mechanisms, as different model structure is found when some components are ablated.
    \item We found known structures (specifically induction heads \citep{elhage2021mathematical}) that were used in unexpected ways. Thus mainline functionality of a component does not always give a full picture.
    \item Finally, we identified heads reliably writing in the \emph{opposite} direction of the correct answer.
\end{compactitem}

\begin{figure*}
\centering
\includegraphics[width=0.8\textwidth]{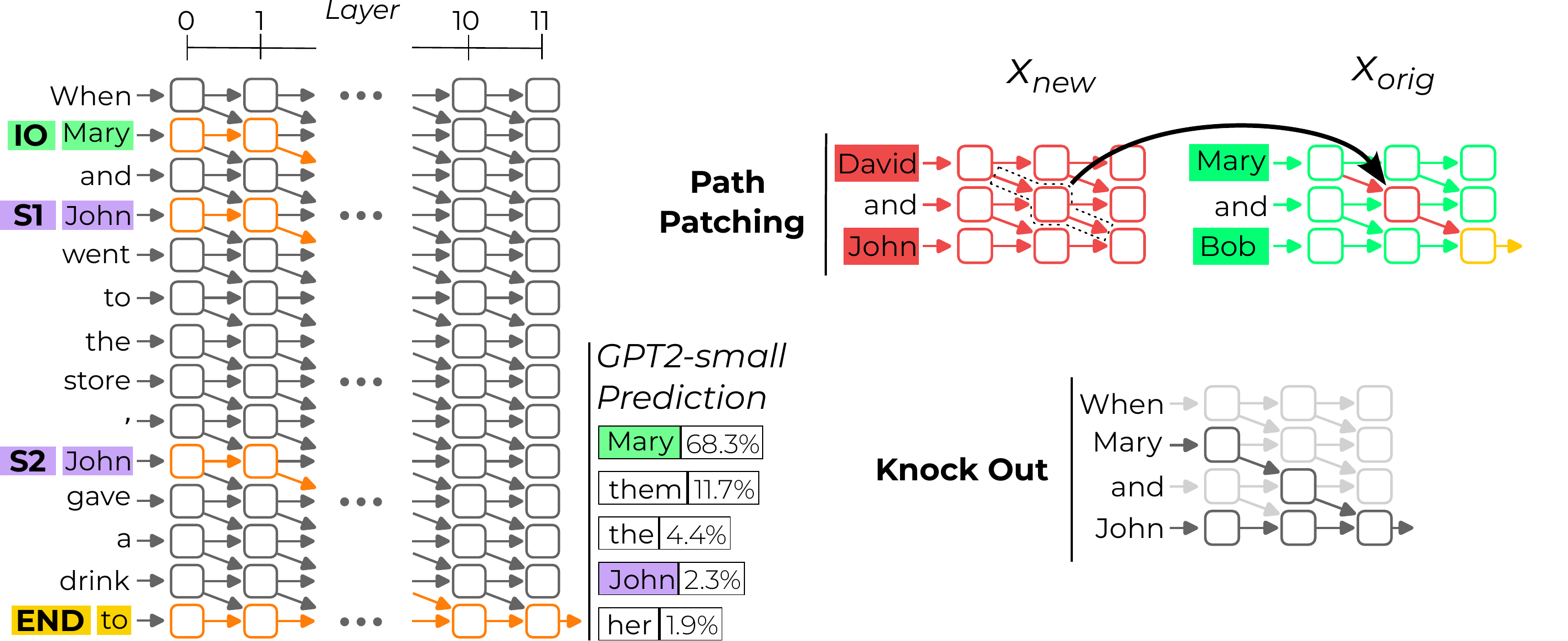}
\caption{Left: We isolated a \emph{circuit} (in orange) responsible for the flow of information connecting the indirect object ``Mary'' to the next token prediction. The nodes are attention layers and the edges represent the interactions between these layers. Right: We discovered and validated this circuit using causal interventions, including both path patching and knockouts of attention heads.}
\label{fig:directed_graph_figure}
\end{figure*}

Explanations for model behavior can easily be misleading or non-rigorous \citep{attention_not_explanation, interp-illusion-bert}. 
To remedy this problem, we formulate three criteria to help validate our circuit explanations. These criteria are \textbf{faithfulness} (the circuit can perform the task as well as the whole model), \textbf{completeness} (the circuit contains all the nodes used to perform the task), and \textbf{minimality} (the circuit doesn't contain nodes irrelevant to the task). 
Our circuit shows significant improvements compared to a na\"{i}ve (but faithful) circuit, but fails to pass the most challenging tests.

In summary, our main contributions are: (1) We identify a large circuit in GPT-2 small that performs indirect-object identification (Figure \ref{fig:the_circuit} and Section \ref{sec:legibility}); (2) Through example, we identify useful techniques for understanding models, as well as surprising pitfalls;  (3) We present criteria that ensure structural correspondence between the circuit and the model, and check experimentally whether our circuit meets this standard (Section \ref{sec:validation}).
\section{Background}
\label{sec:methods}

\label{sec:task_description}
\label{sec:task}

In this section, we introduce the IOI task, review the transformer architecture, define circuits more formally, and describe a technique for ``knocking out'' nodes in a model. 


\textbf{Task description. } A sentence containing indirect object identification (IOI) has an initial dependent clause, e.g ``When Mary and John went to the store'', and a main clause, e.g ``John gave a bottle of milk to Mary''. The initial clause introduces the indirect object (IO) ``Mary'' and the subject (S) ``John''. The main clause refers to the subject a second time, and in all our examples of IOI, the subject gives an object to the IO. The IOI task is to predict the final token in the sentence to be the IO. We use ‘S1’ and ‘S2’ to refer to the first and second occurrences of the subject, when we want to specify position. We create dataset samples for IOI using 15 templates (see Appendix \ref{app:templates}) with random single-token names, places and items. We use the notation $\pioi$ for the distribution over sentences generated by this procedure.

To quantify GPT-2 small's performance on the IOI task, we use two different metrics: logit difference and IO probability. \emph{Logit difference} measures the difference in logit value between the two names, where a positive score means the correct name (IO) has higher probability. \emph{IO probability} measures the absolute probability of the IO token under the model's predictions. Both metrics are averaged over $\pioi$. Over 100,000 dataset examples, GPT-2 small has mean logit difference of 3.56 (IO predicted over S 99.3\% of the time), and mean IO probability of 49\%. 


\textbf{Transformer architecture.} 
GPT-2 small is a decoder-only transformer with 12 layers and 12 attention heads per attention layer. In this work, we mostly focus on understanding the behavior of attention heads, which we describe below using notation similar to \citet{elhage2021mathematical}. We leave a full description of the model to Appendix \ref{app:notation}. 

The input $x_0$ to the transformer is a sum of position and token embeddings and lies in $\mathbb{R}^{N \times d}$, where $N$ is the number of tokens in the input and $d$ is the model dimension. This input embedding is the initial value of the \textit{residual stream}, which all attention layers and MLPs read from and write to. Attention layer $i$ of the network takes as input $x_{i} \in \mathbb{R}^{N \times d}$, the value of the residual stream at layer $i$. The attention layer output can be decomposed into the sum of {attention heads} $h_{i, j}$. If the output of the attention layer is $y_i = \sum_j h_{i, j}(x_i)$, then the residual stream is updated to $ x_i + y_i$.

Focusing on individual heads, each head $h_{i,j}$ is parametrized by four matrices $W_Q^{i,j}$, $W_K^{i,j}$, $W_O^{i,j} \in \mathbb{R}^{d \times \frac{d}{H}}$ and $W_V^{i,j} \in \mathbb{R}^{\frac{d}{H} \times d}$. 
We rewrite these parameters as low-rank matrices in $\mathbb{R}^{d \times d}$: $W_{OV}^{i,j} = W_O^{i,j} W_V^{i,j}$ and $W_{QK}^{i,j} = W_Q^{i,j} (W_K^{i,j})^T$. The QK matrix is used to compute the attention pattern $A_{i, j} \in \mathbb{R}^{N\times N}$ of head $(i, j)$, while the OV matrix determines what is written into the residual stream. At the end of the forward pass, a layer norm is applied before the unembed matrix $W_U$ projects the residual stream into logits.




\subsection{Circuits and Knockouts}
\label{sec:circuits}
\label{sec:knockouts}

In mechanistic interpretability, we want to understand the correspondence between the components of a model and human-understandable concepts. A useful abstraction for this goal is \emph{circuits}. If we think of a model as a computational graph $M$ where nodes are terms in its forward pass (neurons, attention heads, embeddings, etc.) and edges are the interactions between those terms (residual connections, attention, projections, etc.), a circuit $C$ is a subgraph of $M$ responsible for some behavior (such as completing the IOI task). 
This definition of a circuit is more coarse-grained than the one presented in \citet{olah2020zoom}, where nodes are {features} (meaningful directions in the latent space of a model) instead of model components.

Just as the entire model $M$ defines a function $M(x)$ from inputs to logits, we also associate each circuit $C$ with a function $C(x)$ via \emph{knockouts}. A {knockout} removes a set of nodes $K$ in a computational graph $M$ with the goal of ``turning off'' nodes in $K$ but capturing all other computations in $M$. Thus, $C(x)$ is defined by knocking out all nodes in $M \backslash C$ and taking the resulting logit outputs in the modified computational graph.

A first naïve knockout approach consists of simply deleting each node in $K$ from $M$. 
The net effect of this removal is to \textit{zero ablate} $K$, meaning that we set its output to 0. This naïve approach has an important limitation: 0 is an arbitrary value, and subsequent nodes might rely on the average activation value as an implicit bias term. Perhaps because of this, we find zero ablation to lead to noisy results in practice. 

To address this, we instead knockout nodes through \textit{mean ablation}: replacing them with their average activation value across some reference distribution, similar to the bias correction method used in \citet{meca_interp_grokking}.  Mean-ablations remove the information that \textit{varies} in the reference distribution (e.g.~the value of the name outputted by a head) but will preserve \textit{constant} information (e.g.~the fact that a head is outputting a name). 


In this work, all knockouts are performed in a modified $\pioi$ distribution called $\pabc$. It relies on the same generation procedure, but instead of using two names (IO and S) it used three unrelated random names (A, B and C). In $\pabc$, sentences no longer have a single plausible IO, but the grammatical structures from the $\pioi$ templates are preserved.

We chose this distribution for mean-ablating because using $\pioi$ would not remove enough information helpful for the task. For instance, some information constant in $\pioi$ (e.g. the fact that a name is duplicated) is removed when computing the mean on $\pabc$.

When knocking out a single node, a (head, token position) pair in our circuit, we want to preserve the grammatical information unrelated to IOI contained in its activations. However, the grammatical role (subject, verb, conjunction etc.) of a particular token position varies across templates. To ensure that grammatical information is constant when averaging, we compute the mean of a node across samples of the same template.

\begin{figure}[b]
\centering
\includegraphics[width=0.8\textwidth]{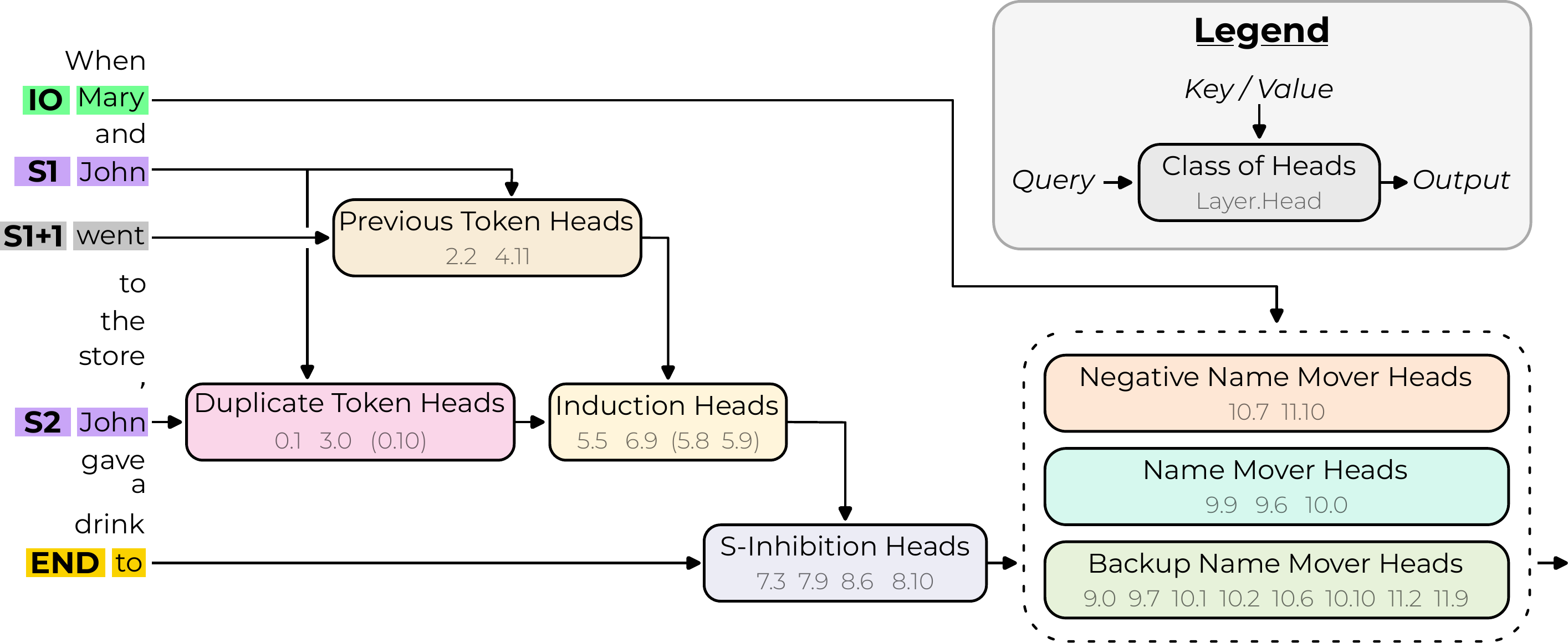}
\caption{We discover a circuit in GPT-2 small that implements IOI. The input tokens on the left are passed into the residual stream. Attention heads move information between residual streams: the query and output arrows show which residual streams they write to, and the key/value arrows show which residual streams they read from.}
\label{fig:the_circuit}
\end{figure}

\section{Discovering the Circuit}
\label{sec:legibility}

\label{sec:overview}

We seek to explain how GPT-2 small implements the IOI task (Section \ref{sec:task_description}). Recall the example sentence ``When Mary and John went to the store, John gave a drink to". The following human-interpretable algorithm suffices to perform this task:
\begin{compactenum}
    \item Identify all previous names in the sentence (Mary, John, John).
    \item Remove all names that are duplicated (in the example above: John).
    \item Output the remaining name.
\end{compactenum}

Below we present a circuit that we claim implements this functionality.
Our circuit contains three major classes of heads, corresponding to the three steps of the algorithm above:
\begin{compactitem}
    \item \emph{Duplicate Token Heads} identify tokens that have already appeared in the sentence. They are active at the S2 token, attend primarily to the S1 token, and signal that token duplication has occurred by writing the position of the duplicate token.
    \item \emph{S-Inhibition Heads} remove duplicate tokens from Name Mover Heads' attention. They are active at the END token, attend to the S2 token, and write in the query of the Name Mover Heads, inhibiting their attention to S1 and S2 tokens.
    \item \emph{Name Mover Heads} output the remaining name. They are active at END, attend to previous names in the sentence, and copy the names they attend to. Due to the S-Inhibition Heads, they attend to the IO token over the S1 and S2 tokens. 
\end{compactitem}
A fourth major family of heads writes in the opposite direction of the Name Mover Heads, thus decreasing the confidence of the predictions. We speculate that these \emph{Negative Name Mover Heads} 
might help the model ``hedge'' so as to avoid high cross-entropy loss when making mistakes. 

There are also three minor classes of heads that perform related functions to the components above:

\begin{compactitem}
    \item \emph{Previous Token Heads} 
    copy information about the token S to token S1+1, the token after S1. 
    \item \emph{Induction Heads} perform the same role as the Duplicate Token Heads through an induction mechanism. They are active at position S2, attend to token S1+1 (mediated by the Previous Token Heads), and 
    their output is used both as a pointer to S1 and as a signal that S is duplicated.
    \item Finally, \emph{Backup Name Mover Heads} do not normally move the IO token to the output, but take on this role if the regular Name Mover Heads are knocked out.
\end{compactitem}

In all of our experiments, we do not intervene on MLPs, 
layer norms,  or embedding matrices, focusing our investigation on understanding the attention heads. 
In initial investigations, we found that knocking out each MLP individually led to good task performance, except for the first layer. However, knocking out all MLPs after the first layer makes the model unable to perform the task (Appendix \ref{app:mlp}). A more precise investigation of the role of MLPs is left for future work.

Below we show how we discovered each of the seven components above, providing evidence that they behave as claimed. We found that it was easiest to uncover the circuit starting at the logits and working back step-by-step. Each step is divided into two parts. First, we trace the information flow backward from previously discovered components by finding attention heads that directly influence them; we do this using a technique we introduce called \emph{path patching}, which generalizes the application of causal mediation analysis to language models introduced in \citet{vig2020investigating}. Second, we characterize the newly identified heads by examining their attention patterns and performing experiments tailored to their hypothesized function.

\subsection{Which heads directly affect the output? (Name Mover Heads)}
\label{sec:name movers} \label{sec:name_movers}
\label{sec:negative heads}

\textbf{Tracing back the information flow.} We begin by searching for attention heads $h$ directly affecting the model's logits. To differentiate indirect effect (where the influence of a component is mediated by another head) from direct effect, we designed a technique called \textbf{path patching} (Figure \ref{fig:directed_graph_figure}). 


Path patching replaces part of a model's forward pass with activations from a different input.
Given inputs $\xorig$ and $\xnew$, and a set of paths $\mathcal{P}$ emanating from a node $h$, path patching runs a forward pass on $\xorig$, but for the paths in $\mathcal{P}$
it replaces the activations for $h$ with those from $\xnew$. In our case, $h$ will be a fixed attention head
and $\mathcal{P}$ consists of all direct paths from $h$ to a set of components $R$, i.e. paths through residual connections
and MLPs (but not through other attention heads); this measures the counterfactual effect of $h$
on the members of $R$. The layers after the element from $R$ are recomputed as in a normal forward pass. See Appendix~\ref{app:path_patching} for full details of the method.  

We will always take $\xorig$ to be a sample from $\pioi$, and $\xnew$ the corresponding
sample from $\pabc$ (i.e.~where the names in the sentence are replaced by three random names). We run
path patching on many random samples from $\pioi$ and measure how this affects the average logit difference. Pathways $h \to R$ that are critical to the model's computation should induce a large drop in logit difference when patched, since the $\pabc$ distribution removes the information needed to complete the IOI task.

\begin{figure}
    \centering
    \hfill
     \begin{subfigure}[b]{0.20\textwidth}
         \centering
         \includegraphics[width=\textwidth]{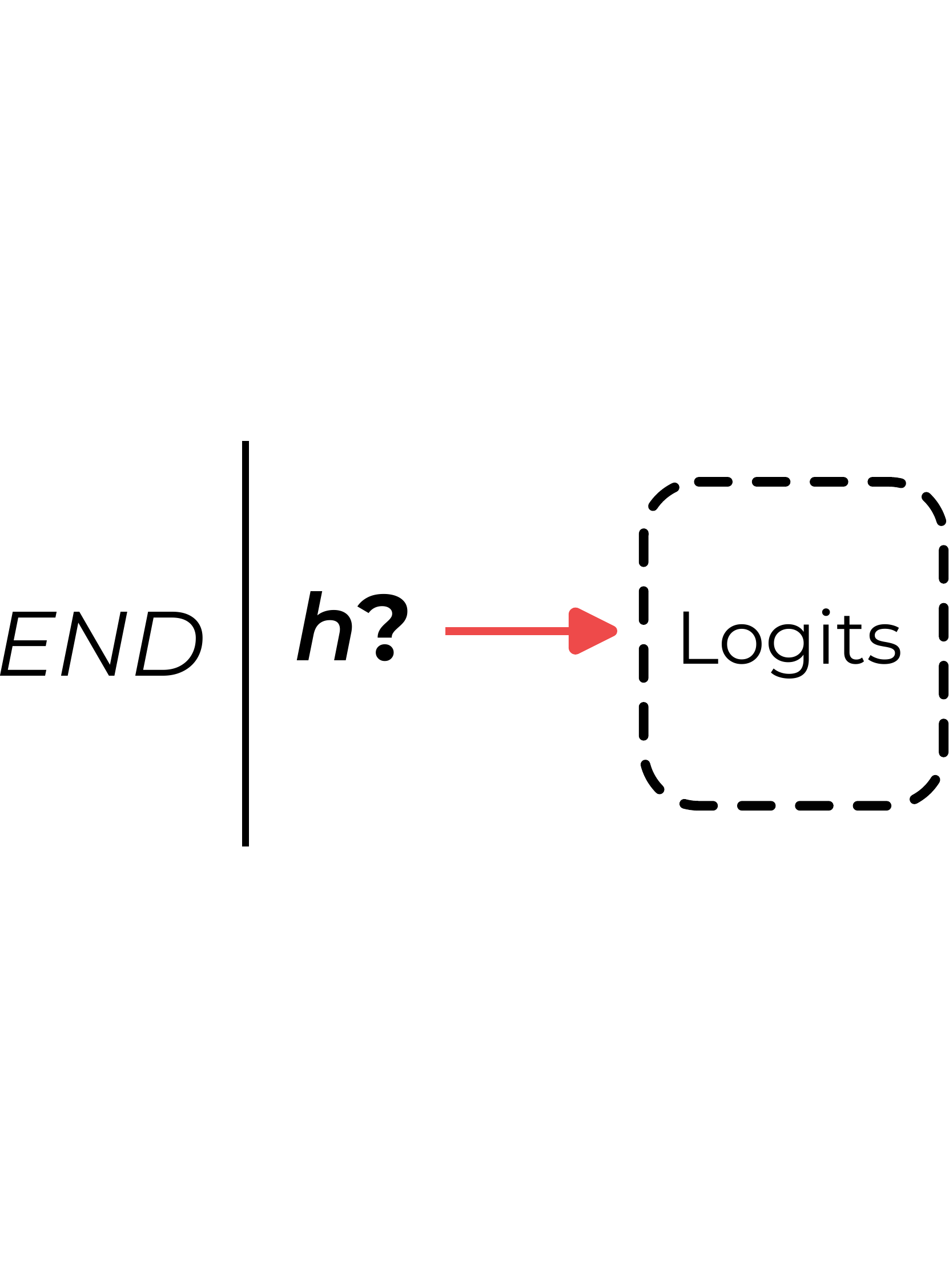}
         \caption{}
         \label{fig:name_moversA}
     \end{subfigure}
     \hfill
     \begin{subfigure}[b]{0.30\textwidth}
         \centering
         \includegraphics[width=\textwidth]{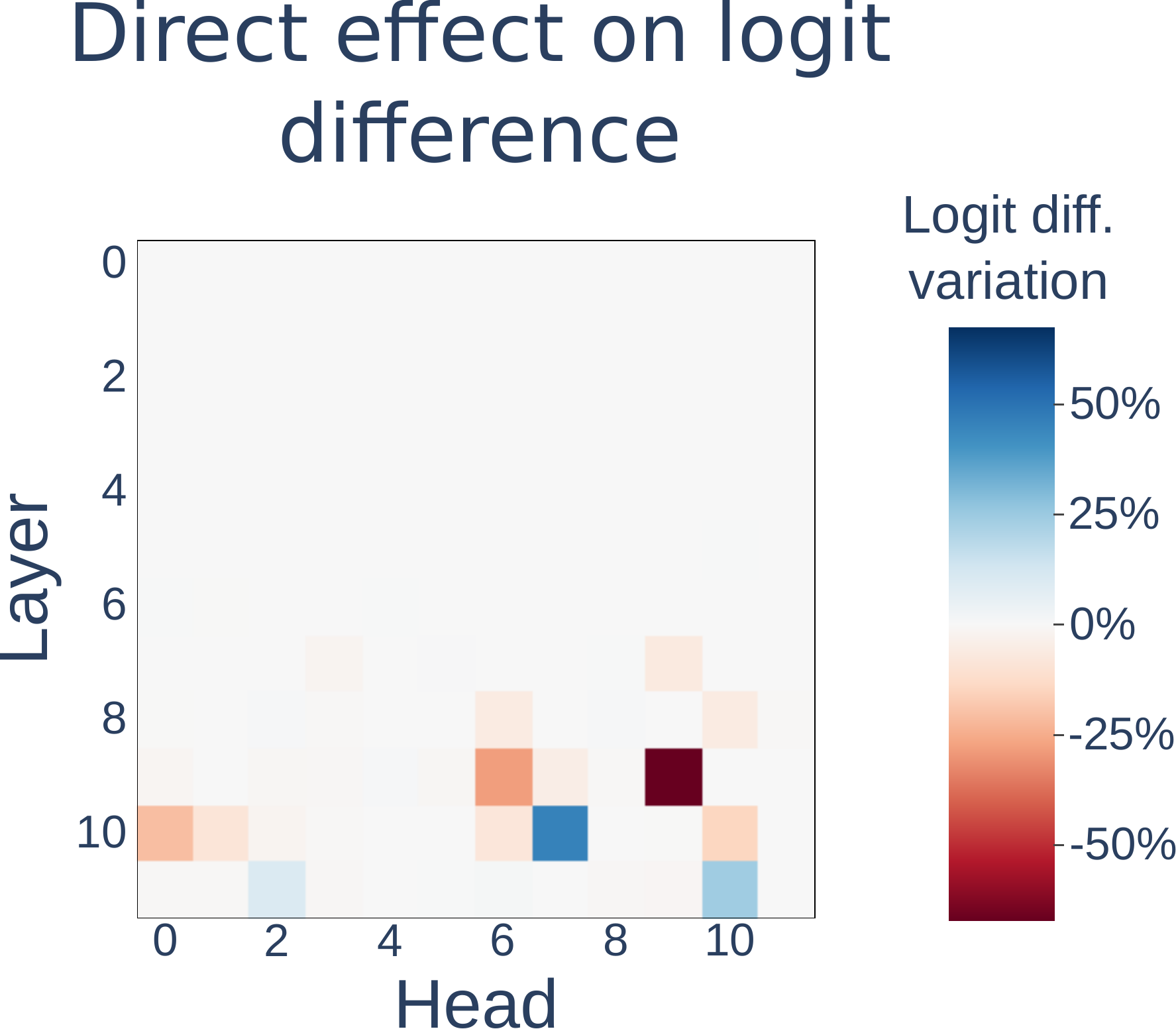}
         \caption{}
         \label{fig:name_moversB}
     \end{subfigure}
     \hfill
     \begin{subfigure}[b]{0.40\textwidth}
         \centering
         \includegraphics[width=\textwidth]{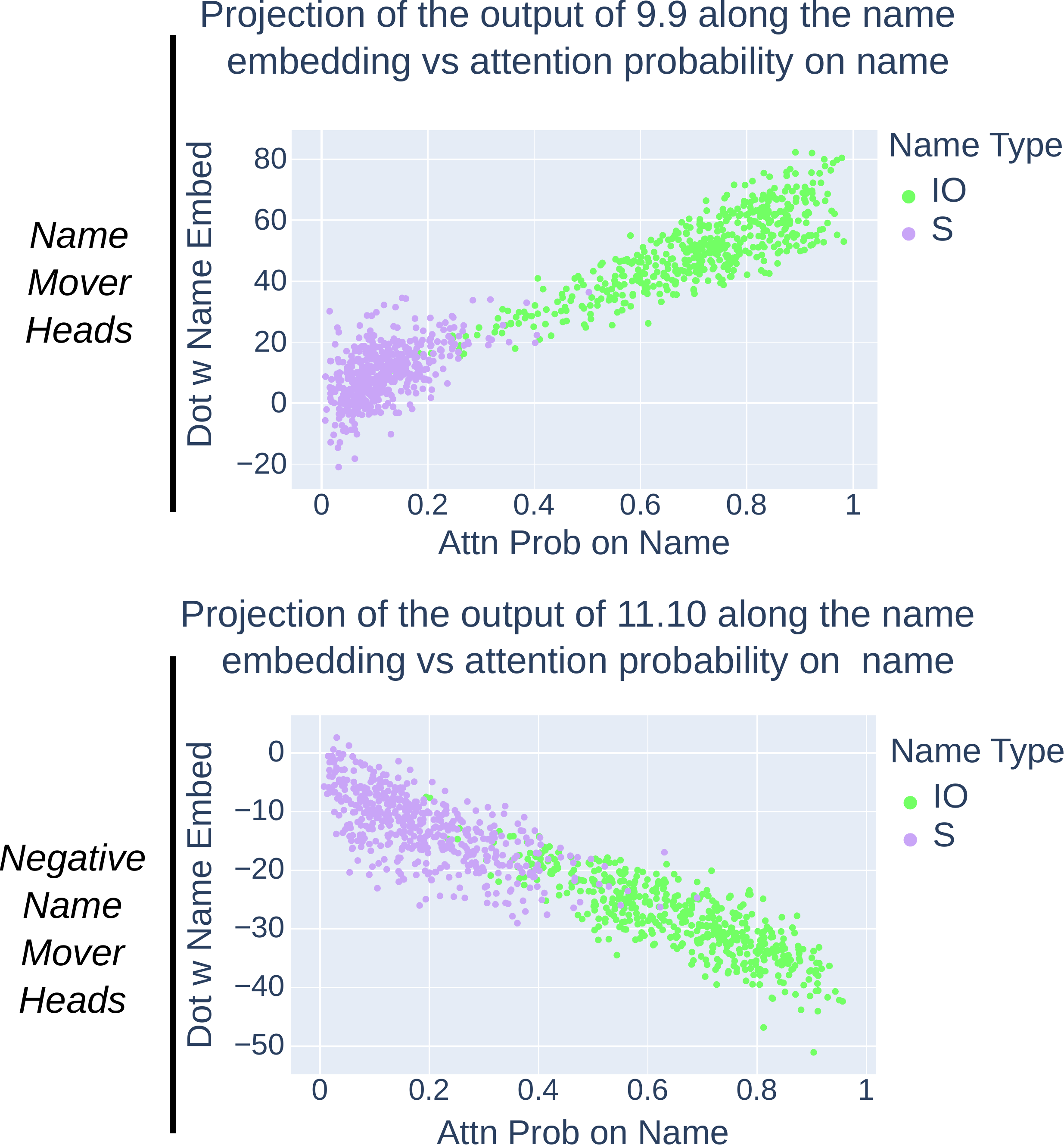}
         \caption{}
         \label{fig:name_moversC}
     \end{subfigure}
     \hfill
    \caption{(a) We are searching for heads $h$ directly affecting the logits using path patching. (b) Results of the path patching experiments. Name Movers and Negative Name Movers Heads are the heads that have the strongest direct effect on the logit difference. (c) Attention probability vs projection of the head output along $W_{U}[IO]$ or $W_{U}[S]$ respectively. For S tokens, we sum the attention probability on both S1 and S2. }
    \label{fig:name_movers}
\end{figure}

To trace information back from the logits, we run path patching for the pathway $h \to \text{Logits}$ for each head $h$ at the END position and display the effect on logit difference in Figure \ref{fig:name_moversA}.
We see that only a few heads in the final layers cause a large effect on logit difference. Specifically, patching 9.6, 9.9, and 10.0 causes a large drop (they thus contribute positively to the logit difference), while 10.7 and 11.10 cause a large increase (they contribute negatively to the logit difference).

\textbf{Name Mover Heads.} To understand the heads positively influencing the logit difference, we first study their attention patterns. We find that they attend strongly to the IO token: the average attention probability of all heads over $\pioi$ is 0.59. We hypothesize that these heads (i) attend to the correct name and (ii) copy whatever they attend to. We therefore call these heads \emph{Name Mover Heads}. 


To verify our hypothesis, we design experiments to test the heads' functionality.
Let $W_U$ denote the unembedding matrix, and $W_U[IO]$, $W_U[S]$ the corresponding unembedding vectors for the $IO$ and $S$ tokens. We scatter plot the attention probability against the logit score $\langle h_i(X), W_U[N]  \rangle$, measuring how much head $h_i$ on input $X$ is writing in the direction of the logit of the name $N$ (IO or S). The results are shown in Figure~\ref{fig:name_moversC}: higher attention probability on the IO or S token is correlated with higher output in the direction of the name (correlation $\rho>0.81$, $N=500$)\footnote{In GPT-2 small, there is a layer norm before the final unembedding matrix. This means that these dot products computed are not appropriately scaled. However, empirically we found that approaches to approximating the composition of attention head outputs with this layer norm were complicated, and resulted in similar correlation and scatter plots.}. 

To check that the Name Mover Heads copy names generally, we studied what values are written via the heads' OV matrix. Specifically, we first obtained the state of the residual stream at the position of each name token after the first MLP layer. Then, we multiplied this by the OV matrix of a Name Mover Head (simulating what would happen if the head attended perfectly to that token), multiplied by the unembedding matrix, and applied the final layer norm to obtain logit probabilities. We compute the proportion of samples that contain the input name token in the top 5 logits ($N = 1000$) and call this the copy score. All three Name Mover Heads have a copy score above $95\%$, compared to less than $20\%$ for an average head.


\textbf{Negative Name Mover Heads.}
In Figure \ref{fig:name_moversB}, we also observed two heads causing a large increase, and thus negatively influencing the logit difference. We called these heads \emph{Negative Name Mover Heads}. These share all the same properties as Name Mover Heads except they (1) write in the opposite direction of names they attend to and (2) have a large \textit{negative} copy score--the copy score calculated with the negative of the OV matrix (98\% compared to 12\% for an average head).  


\subsection{Which heads affect the Name Mover Heads' attention? (S-Inhibition Heads)}
\label{sec:s_inhibition}

\begin{figure}
    \centering
     \begin{subfigure}[b]{0.3\textwidth}
         \centering
         \includegraphics[width=\textwidth]{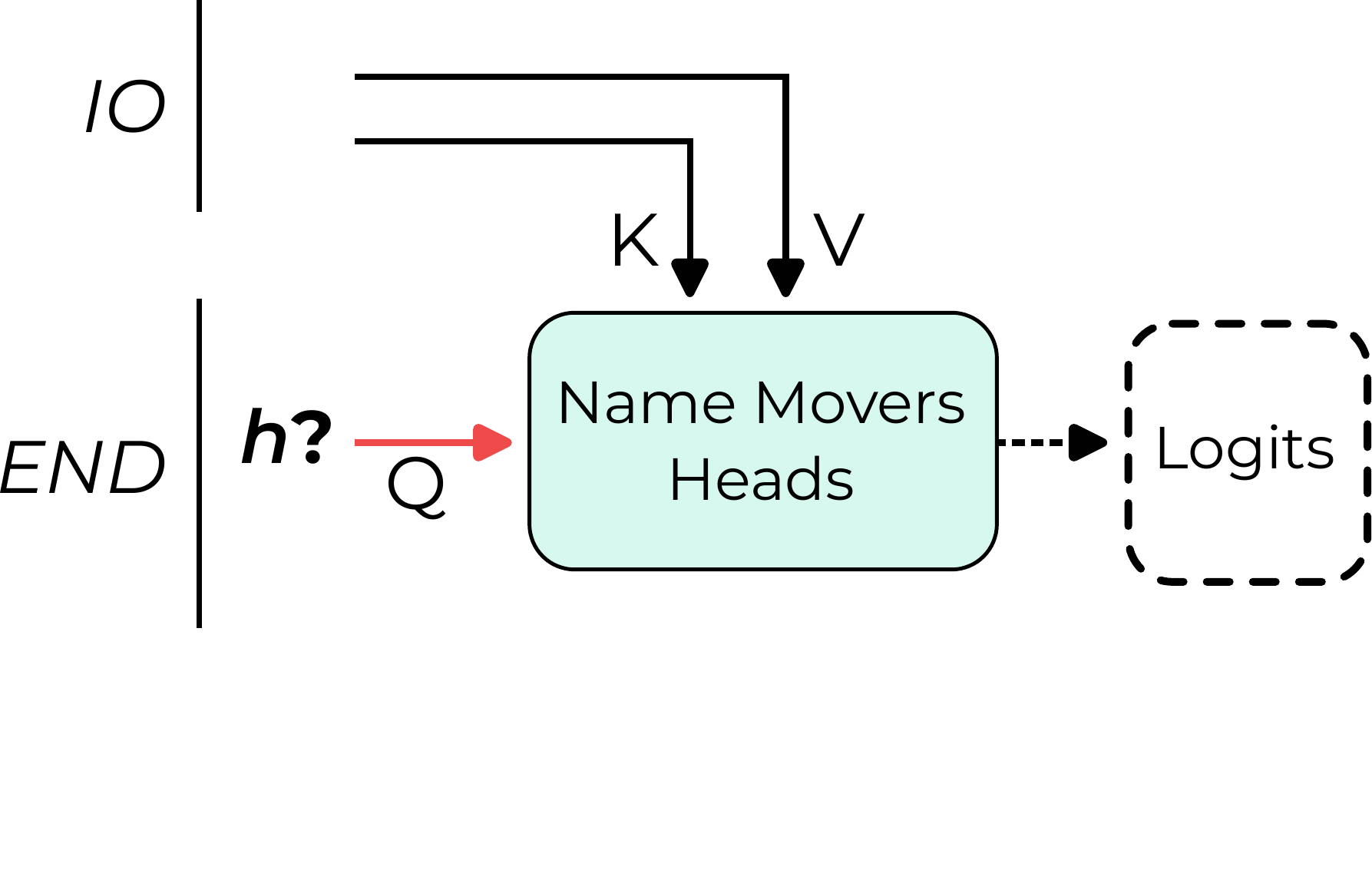}
         \caption{}
         \label{fig:sinA}
     \end{subfigure}
     \hfill
     \begin{subfigure}[b]{0.30\textwidth}
         \centering
         \includegraphics[width=\textwidth]{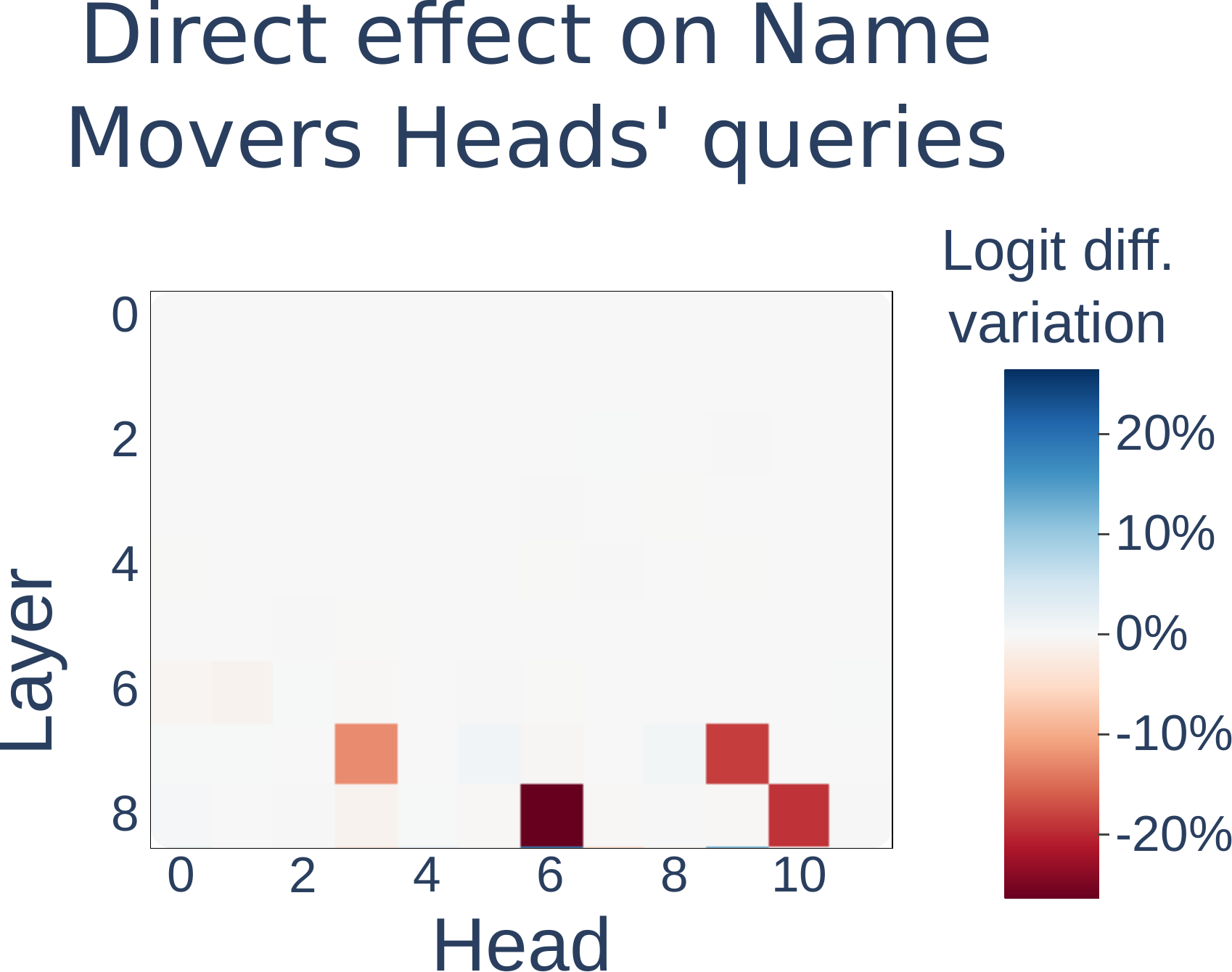}
         \caption{}
         \label{fig:sinB}
     \end{subfigure}
     \hfill
     \begin{subfigure}[b]{0.35\textwidth}
         \centering
         \includegraphics[width=\textwidth]{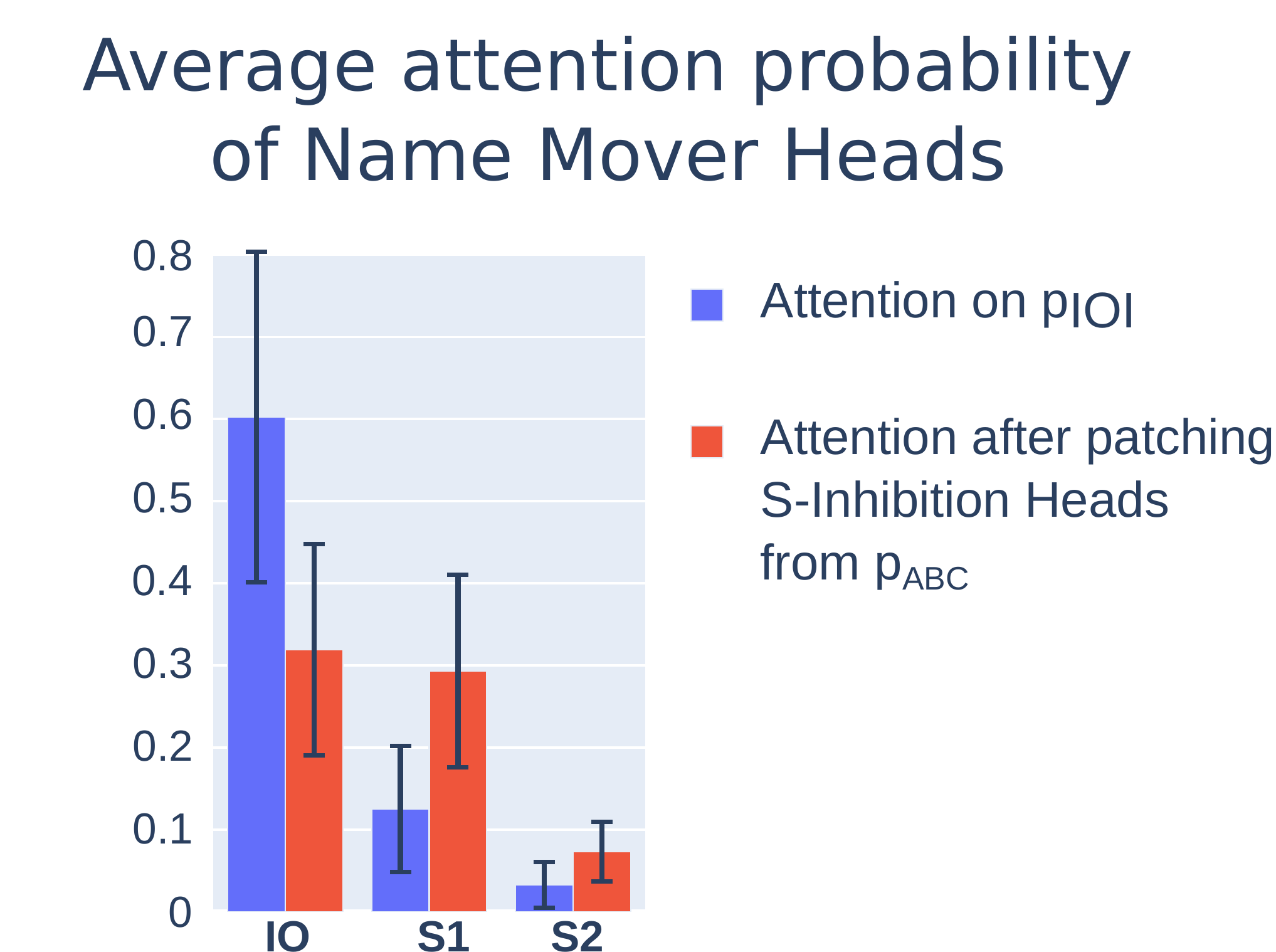}
         \caption{}
         \label{fig:sinC}
     \end{subfigure}
    \caption{(a) Diagram of the direct effect experiments on Name Mover Heads' queries. We patch in the red path from the $\pabc$ distribution. Then, we measure the indirect effect of $h$ on the logits (dotted line). All attention heads are recomputed on this path. (b) Result of the path patching experiments. The four heads causing a decrease in logit difference are S-Inhibition Heads. (c) Attention of Name Mover Heads on $\pioi$ before and after path patching of S-Inhibition Heads $\to$ Name Mover Heads' queries for all four S-Inhibition Heads at once (black bars show the standard deviation). S-Inhibition Heads are responsible for Name Mover Heads' selective attention on IO.}
    \label{fig:sin}
\end{figure}

\textbf{Tracing back the information flow.} Given that the Name Mover Heads are primarily responsible for constructing the output, we next ask what heads they depend on. As illustrated in Figure~\ref{fig:sinA}, there are three ways to influence attention heads: by their values, keys, or queries. In the case of Name Mover Heads, their value matrix appears to copy the input tokens (see previous section), so we do not study it further. Moreover, as the IO token appears early in the context, the Name Mover Heads' key vectors for this position likely do not contain much task-specific information. Therefore, we focus on the query vector, which is located at the END position. 

We again use path patching, identifying the heads that affect each Name Mover Head's query vector. Specifically, for a head $h$ we patch the path $h \to \text{Name Mover Heads' queries}$ (Figure~\ref{fig:sinA}) and report the effect on logit difference in Figure~\ref{fig:sinB}. 
%
Heads 7.3, 7.9, 8.6, 8.10 directly influence the Name Mover Heads' queries, causing a significant drop in logit difference through this path. 

\textbf{S-Inhibition Heads.} To better understand the influence of these heads on the Name Movers Heads, we visualized Name Movers Heads' attention before and after applying path patching of all four previously identified heads at once (Figure \ref{fig:sinC}). Patching these heads leads to a greater probability on S1 (and thus less on IO). We therefore call these heads \emph{S-Inhibition Heads}, since their effect is to decrease attention to S1.

\begin{figure}
    \centering
     \begin{subfigure}[b]{0.43\textwidth}
         \centering
         \includegraphics[width=\textwidth]{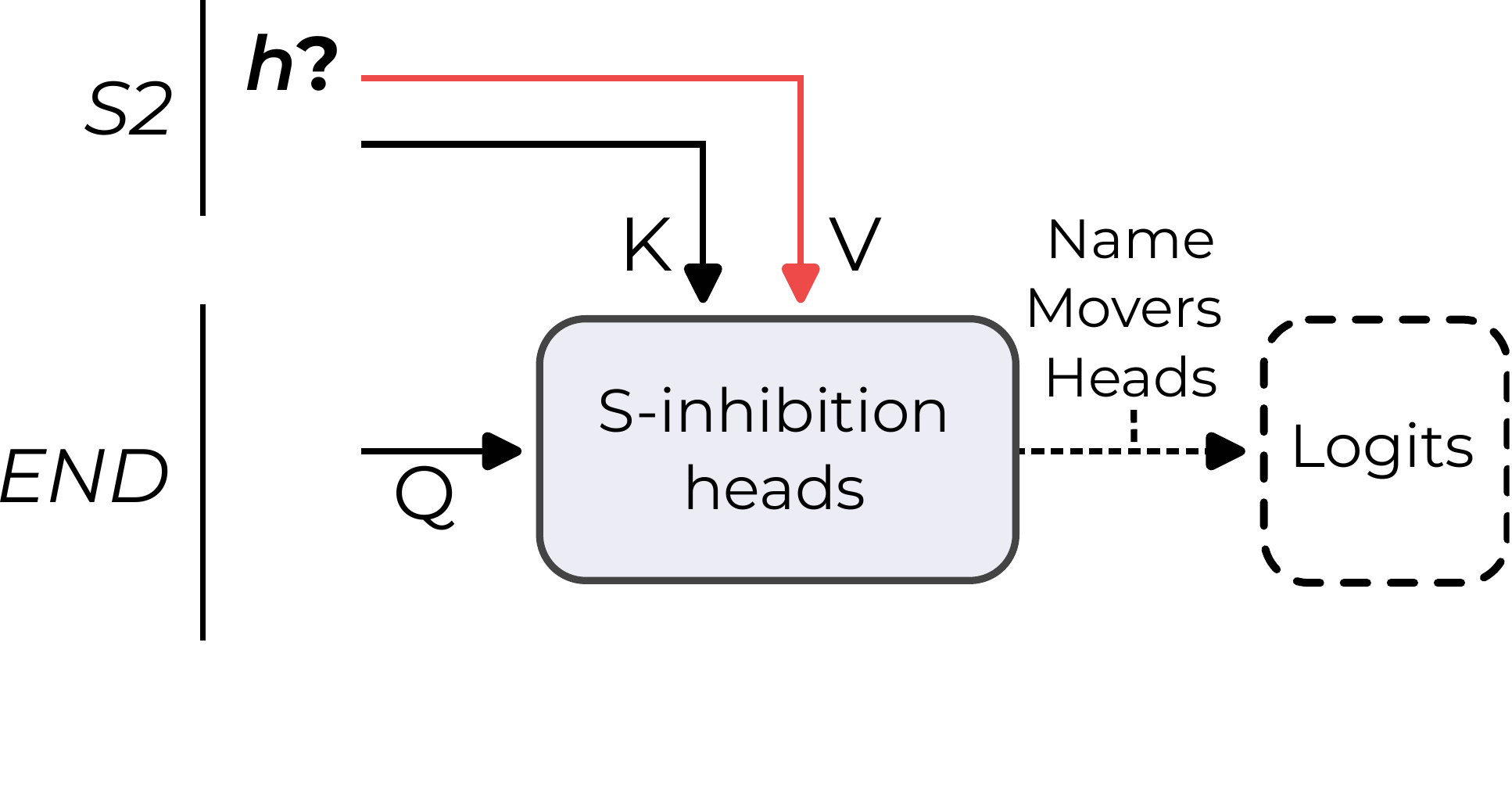}
         \caption{}
         \label{fig:effect_on_sinA}
     \end{subfigure}
     \hfill
     \begin{subfigure}[b]{0.43\textwidth}
         \centering
         \includegraphics[width=\textwidth]{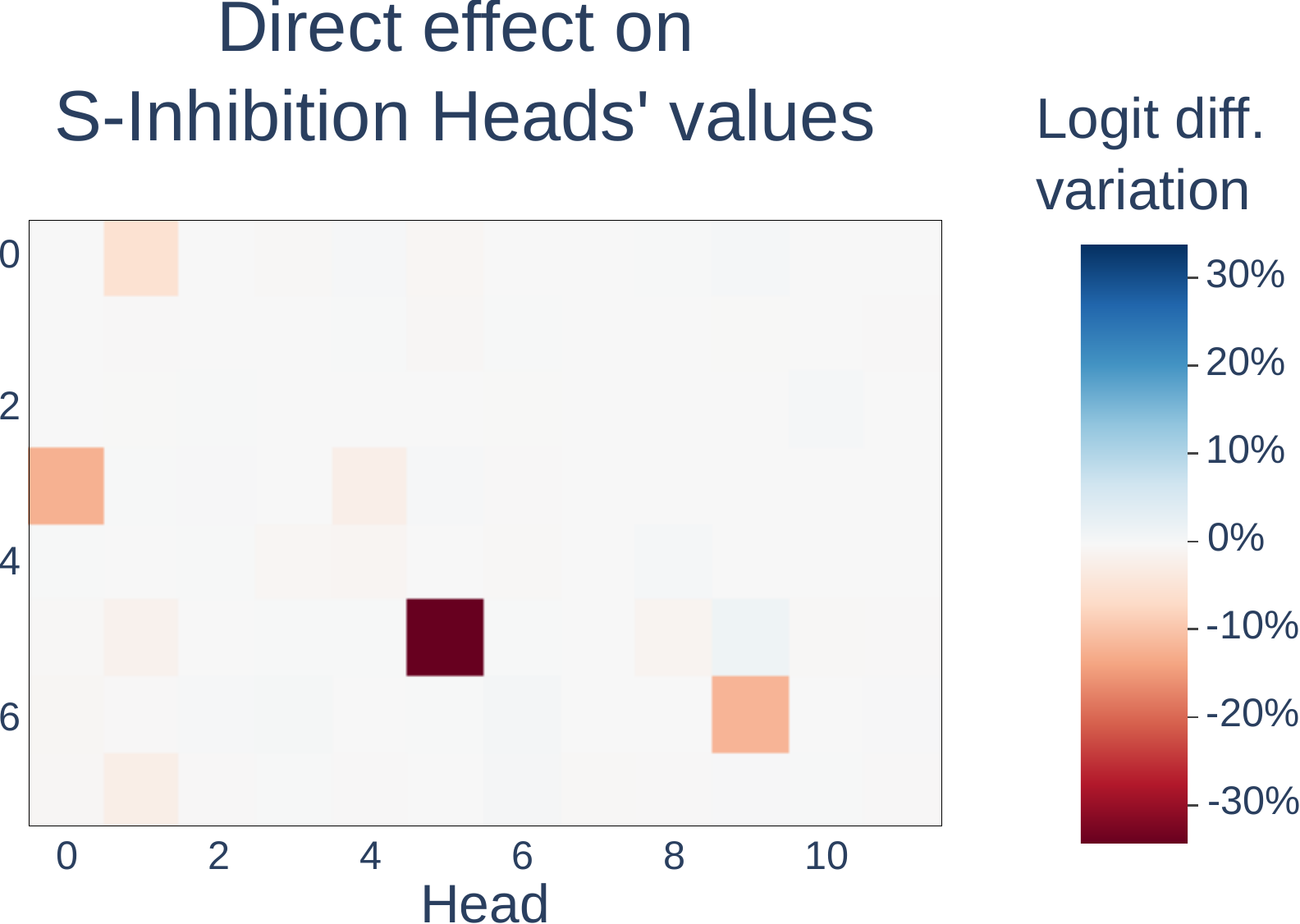}
         \caption{}
         \label{fig:effect_on_sinB}
     \end{subfigure}
    \caption{(a) Diagram of the direct effect experiment on S-Inhibition Heads' values. On the indirect path from $h$ to the logits (dotted line), all elements are recomputed, Name Movers Heads are mediating the effect on logits. (b) Result of the path patching experiments for the S-Inhibition Heads' values.}
    \label{fig:effect_on_sin}
\end{figure}


\textbf{What are S-Inhibition heads writing?}
We observed that S-Inhibition heads preferentially attend to the S2 token (the average attention from END to S2 was 0.51 over the four heads). We conjectured that they move information about S that causes Name Movers to attend less to S1. We discovered two signals through which S-Inhibition Heads do this. The first, the \emph{token signal}, contains the value of the token S and causes Name Mover Heads to avoid occurrences of that token. The second, the \emph{position signal}, contains information about the position of the S1 token, causing Name Mover Heads to avoid the S1 position no matter the value of the token at this position.

To disentangle the two effects, we designed a series of counterfactual experiments where only some signals are present, and the others are inverted with respect to the original dataset (see Appendix~\ref{app:disantangling_signals}). 
Both effects matter, but position signals have a greater effect than token signals. These two signals suffice to account for the full average difference in Name Mover Heads' attention to IO over S1.

S-Inhibition heads output information about the position of S1 but don't attend to it directly. They thus must rely on the output of other heads through value composition, as we will see next.

\subsection{Which heads affect S-Inhibition values?}
\label{sec:s_inhibition2}

\textbf{Tracing back the information flow.} We next trace the information flow back through the queries, keys, and values of the S-Inhibition Heads, using path patching as before. We found the values were the most important and so focus on them here, with keys discussed in Appendix~\ref{app:s_in_keys} (queries had no significant direct effect on the S-Inhibition Heads).

In more detail, for each head $h$ at the S2 position we patched in the path $h \to \text{S-Inhibition Heads's values}$ (Figure~\ref{fig:effect_on_sinA}) and reported results in Figure~\ref{fig:effect_on_sinB}. 

We identified 4 heads with a significant negative effect on logit difference.
By looking at their attention pattern, we separated them into two groups. The heads from the first group attend from S2 to S1, while those from the second group attend from S2 to the token after S1 (S1+1 in short). 

\textbf{Duplicate Token Heads.} Two of the heads attend from S2 to S1. We hypothesize that these heads attend to the previous occurrence of a duplicate token and copy the position of this previous occurrence to the current position. We therefore call them \emph{Duplicate Token Heads}.


To partially validate this hypothesis, we investigated these heads' attention patterns 
on sequences of random tokens (with no semantic meaning). We found that the 2 Duplicate Token Heads pay strong attention to a previous occurrence of the current token when it exists (see Appendix \ref{app:validation_dups}). 
%



\textbf{Induction Heads and Previous Token Heads.} The other group of two heads attends from S2 to S1+1. This matches the attention pattern of \emph{Induction Heads} \citep{elhage2021mathematical}, which recognize the general pattern \texttt{[A]\,[B]\,...\,[A]} and contribute to predicting \texttt{[B]} as the next token. For this, they act in pairs with Previous Token Heads. The Previous Token Heads write information about \texttt{[A]} into the residual stream at \texttt{[B]}. Induction Heads can then match the next occurrence of \texttt{[A]} to that position (and subsequently copy \texttt{[B]} to the output).

To test the hypothesis that this group of heads consists of Induction Heads, we sought out the Previous Token Heads that they rely on through key composition. To this end, we apply path patching to find heads that influence logit difference through the S1+1 keys of the Induction Heads (see Appendix \ref{app:prev_token_heads}). We identified two heads with significant effect sizes. 
To test the induction mechanism, we followed the method introduced in \citet{olsson2022context} by checking for prefix-matching (attending to \texttt{[B]} on pattern like \texttt{[A]\,[B]\,...\,[A]}) and copying (increasing the logit of the \texttt{[B]} token) of Induction Heads on repeated random sequences of tokens. We also analyzed the attention patterns of Previous Token Heads on the same dataset. As reported in Appendix~\ref{app:random_tok_seq}, Previous Token Heads attend primarily to the previous token and 2 Induction Heads demonstrate both prefix-matching and copying.




\textbf{Locating the position signal.} Despite their differences, the attention patterns of both groups of heads are determined by the position of S1, suggesting that their output could also contain this information. We thus hypothesized that Duplicate Token Heads and Induction Heads are the origin of the position signal described in Section~\ref{sec:s_inhibition}, which is then copied by S-Inhibition Heads. 

To test this hypothesis, we applied the same technique from the previous section to isolate token and position signals. We patched the output of the Induction and Duplicate Token Heads from sentences with altered position signals. We observed a drop in logit difference of almost the same size as patching the S-Inhibition Heads (at least 88\%). However, patching their output from a dataset with a different S1 and S2 token but at the same position doesn’t change the logit difference (less than 8\% of the drop induced by patching S-Inhibition Heads).

\textbf{Missing validations and caveats.} To fully validate the claimed function of the Duplicate Token and Induction Heads, we would want to perform additional checks that we omitted due to time constraints. First, for Duplicate Token Heads, we would want to study the interaction of the QK matrix with the token embeddings to test if it acts as a collision detector. Moreover, we would investigate the OV matrix to ensure it is copying information from positional embeddings.  For Induction Heads, we would perform the more thorough parameter analysis in \cite{elhage2021mathematical} to validate the copying and key composition properties at the parameter level.

Moreover, in the context we study, Induction Heads perform a different function compared to what was described in their original discovery. In our circuit, Induction Heads' outputs are used as positional signals for S-Inhibition values. We did not investigate how this mechanism was implemented. More work is needed to precisely characterize these newly discovered heads and understand how they differ from the original description of Induction Heads.

\subsection{Did we miss anything? The Story of the Backup Name Movers Heads}
\label{sec:distributed_name_movers}


Each type of head in our circuit has many copies, suggesting that the model implements redundant behavior\footnote{It is unlikely that SGD would select for perfectly redundant modules, but redundancy is more likely when analyzing a sub-distribution of all of a model's training data or a subtask of its behavior.}. To make sure that we didn't miss any copies, we knocked out all the Name Mover Heads at once. To our surprise, the circuit still worked (only 5\% drop in logit difference). After the knockout, we found the new heads directly affecting the logits using the same path patching experiment from Section~\ref{sec:name_movers}. 
These new heads compensate for the loss of Name Movers Heads by replacing their role.

We selected the eight heads with the highest direct influence on logit difference after knockout and call them \emph{Backup Name Mover Heads}. We investigated their behavior before the knockout. We observe diverse behavior: 4 heads show a close resemblance to Name Mover Heads, 2 heads equally attend to IO and S and copy them, 1 head pays more attention to S1 and copies it, and 1 head seems to track and copy subjects of clauses, copying S2 in this case. See Appendix \ref{app:backup_name_movers} for further details on these heads.

We hypothesize that this compensation phenomenon is caused by the use of dropout during training. Thus, the model was optimized for robustness to dysfunctional parts. More work is needed to determine the origin of this phenomenon and if such compensation effects can be observed beyond our special case. 

\section{Experimental validation}
\label{sec:validation}

In the previous section, we provided an end-to-end explanation of how the model produces a high logit difference between IO and S on the IOI task. Yet Section \ref{sec:distributed_name_movers} shows that our experimental methods do not fully account for all name-moving in the model: some new heads take on this role after the main Name Mover Heads are knocked out. We therefore seek more systematic validations to check that our account in Section~\ref{sec:circuits} includes all components used by the model to perform IOI.

To formalize this, we first introduce notation to measure the performance of a circuit $C$ on the IOI task. 
Suppose $X\sim \pioi$, and $f(C(X); X)$ is the logit difference between the IO and S tokens when the circuit $C$ is run on the input $X$ (recall that $C(X)$ is defined by knocking out all nodes outside $C$, as described in Section~\ref{sec:knockouts}). We define the average logit difference $F(C) \eqdef \mathbb{E}_{X \sim \pioi} \left[f(C(X); X) \right]$ as a measure of how well $C$ performs the IOI task.

%
Using $F$, we introduce three criteria for validating a circuit $C$. The first one, \textbf{faithfulness}, asks that $C$ has similar performance to the full model $M$, as measured by $|F(M) - F(C)|$. For the circuit $C$ from Figure~\ref{fig:the_circuit}, we find that $|F(M)-F(C)| = 0.46$, or only $13\%$ of $F(M)=3.56$. In other words, $C$ achieves $87\%$ of the performance of $M$.

However, as the Backup Name Mover Heads illustrate, faithfulness alone is not enough to ensure a complete account. In Section~\ref{sec:completeness} we introduce a toy example that illustrates this issue in more detail and use this to define two other criteria---completeness and minimality---for checking our circuit.



\subsection{Completeness}
\label{sec:completeness}


As a running example, suppose a model $M$ uses two identical and disjoint serial circuits $C_1$ and $C_2$. The two sub-circuits are run in parallel before applying an OR operation to their results. Identifying only one of the circuits is enough to achieve faithfulness, but we want explanations that include both $C_1$ and $C_2$, since these are both used in the model's computation.


To solve this problem, we introduce the \textbf{completeness} criterion: for every subset $K \subseteq C$, the \emph{incompleteness score} $|F(C \setminus K) - F(M\setminus K)|$ should be small. In other words, $C$ and $M$ should not just be similar, but remain similar under knockouts. 

In our running example, we can show that  $C_1$ is not complete by setting $K=C_1$. Then $C_1 \setminus K$ is the empty circuit while $M\setminus K$ still contains $C_2$. The metric $|F(C_1 \setminus K)- F(M \setminus K)|$ will be large because $C_1 \setminus K$ has trivial performance while $M \setminus K$ successfully performs the task.

\begin{figure}
    \centering 
     \begin{subfigure}[b]{0.39\textwidth}
         \centering
         \includegraphics[width=\textwidth]{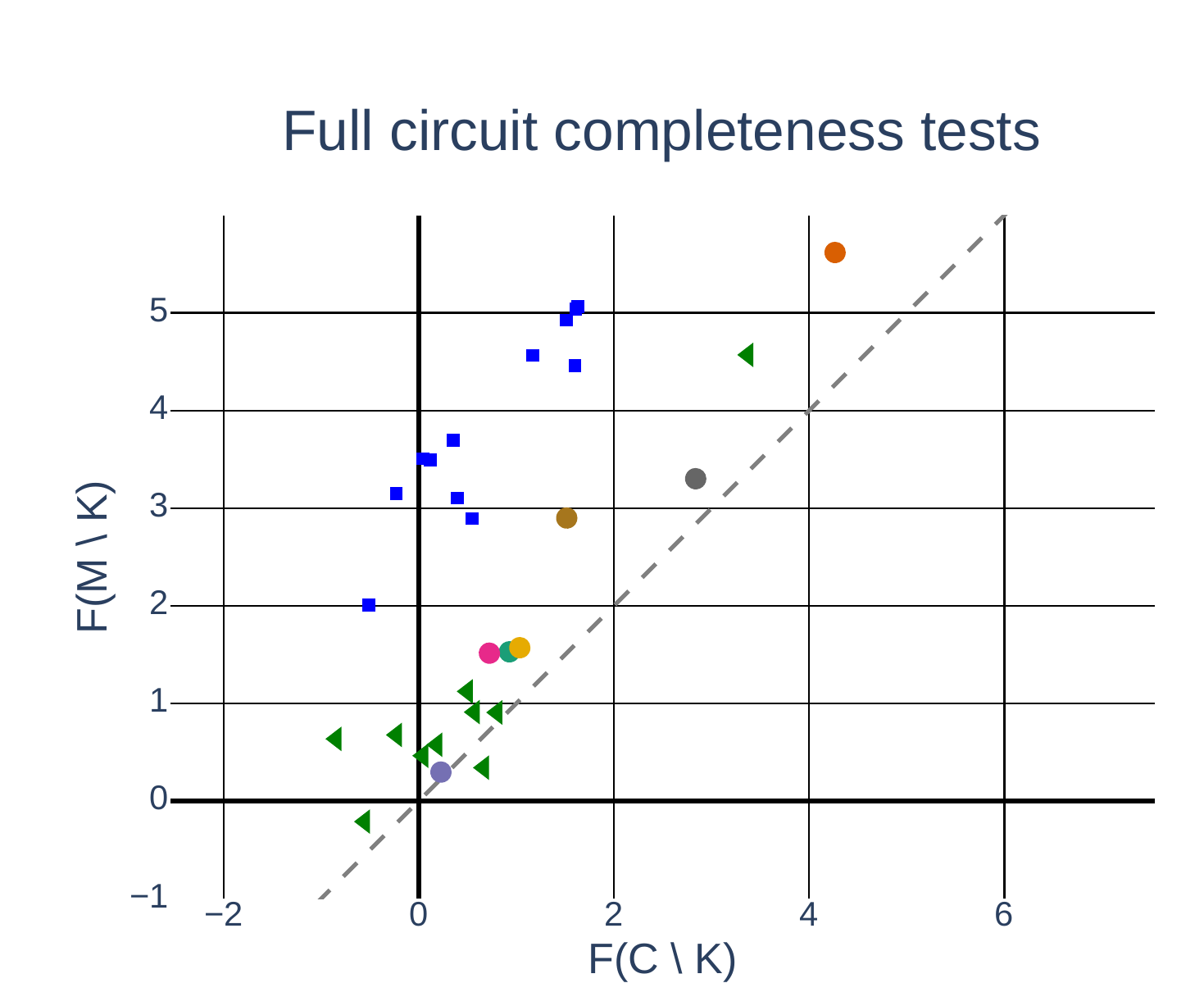}
         \caption{}
         
         \label{fig:completenessA}
     \end{subfigure}
     \hfill
     \begin{subfigure}[b]{0.51\textwidth}
         \centering
         \includegraphics[width=\textwidth]{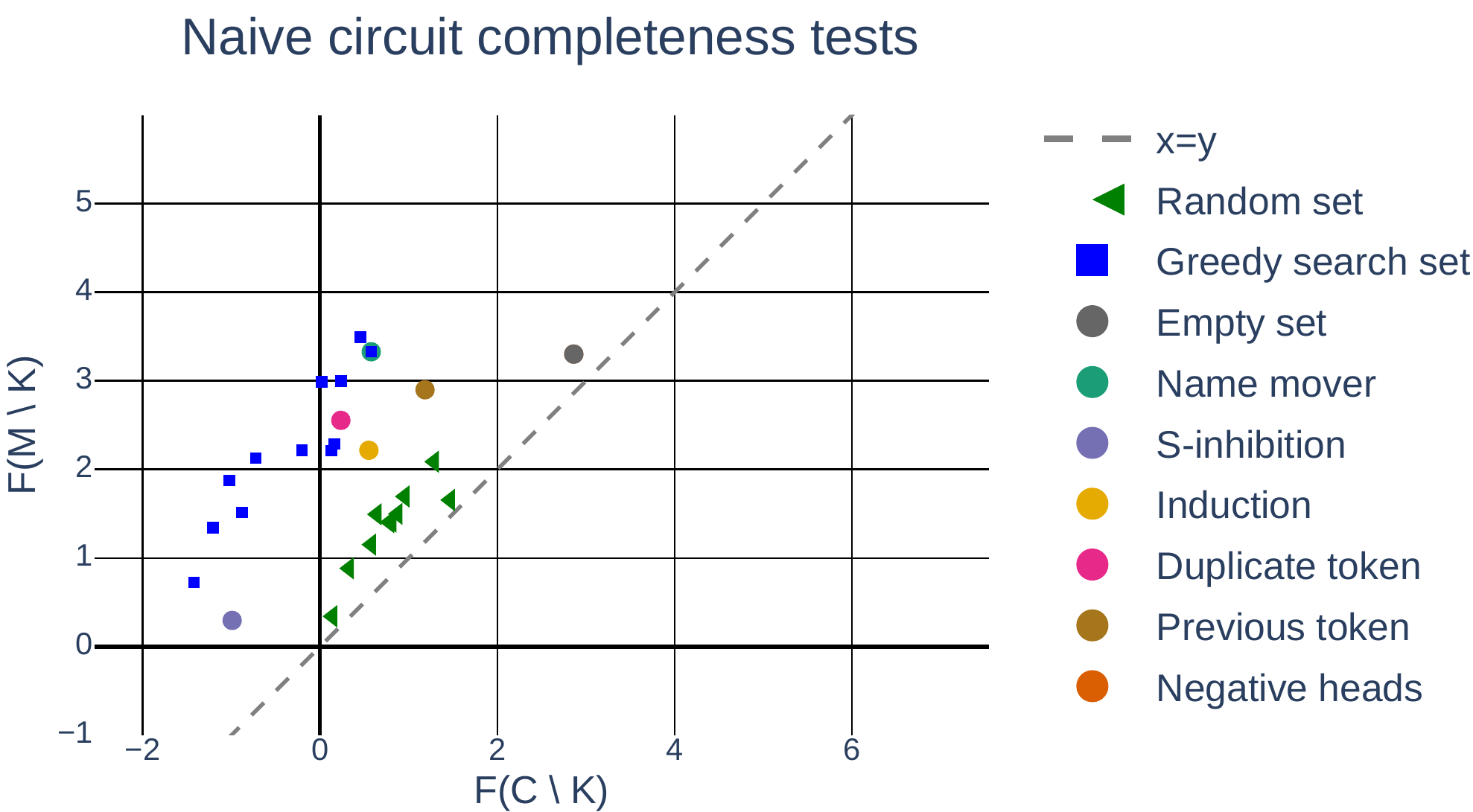}
         \caption{}
         \label{fig:completenessB}
     \end{subfigure}
    \caption{Plot of points $(x_K, y_K) = (F(C \setminus K), F(M \setminus K))$ for (a) our circuit and (b) a naive circuit. Each point is for a different choice of $K$. We show sets $K$ obtained from different sampling strategies: 10 uniformly randomly chosen $K \subseteq C$, $K = \emptyset$, $K$ a class of heads from the circuit, and 10 $K$ found by greedy optimization. Since the incompleteness score is $|x_K - y_K|$, we show the line $y=x$ for reference.}
    \label{fig:completeness}
\end{figure}

Testing completeness requires a search over exponentially many subsets $K \subseteq C$. This is computationally intractable given the size of our circuit, hence we use three sampling methods to seek sets $K$ that give large incompleteness scores:
\begin{compactitem}
\item The first sampling method chooses subsets $K \subseteq C$ uniformly at random.
\item The second sampling method set $K$ to be an entire class of circuit heads $G$, e.g the Name Mover Heads. $C \setminus G$ should have low performance since it's missing a key component, whereas $M \setminus G$ might still do well if it has redundant components that fill in for $G$. 
\item Thirdly, we greedily optimized $K$ node-by-node to maximize the incompleteness score (see Appendix \ref{app:greedy} for details).
\end{compactitem}
All results are shown in Figure \ref{fig:completenessA}.
The first two methods of sampling $K$ suggested to us that our circuit was complete, as every incompleteness score computed with those methods was small.  However, the third resulted in sets $K$ that had high incompleteness score: up to 3.09 (87\% of the original logit difference). These greedily-found sets were usually not semantically interpretable (containing heads from multiple categories), and investigating them would be an interesting direction of future work.




\subsection{Minimality}
\label{sec:minimality}

A faithful and complete circuit may contain unnecessary components, and so be overly complex. 
To avoid this, we should check that each of its nodes $v$ is actually necessary. 
This can be evaluated by showing that $v$ can significantly recover $F$ after knocking out a set of nodes $K$. 

Formally, the \textbf{minimality} criterion is whether for every node $v\in C$ there exists a subset $K \subseteq C \setminus \{v\}$ that has high \emph{minimality score} $|F(C \setminus (K \cup \{ v \})) - F(C \setminus K)| $. 


In the running example, we can show that $C_1 \cup C_2$ is minimal. The proof has two symmetrical cases: $v \in C_1 $ and $v\in C_2$. Without loss of generality, we chose $v \in C_1$, and then define $K = C_2$. The minimality score for this choice of node $v$ and set $K$ is equal to $|F(C_1 \setminus \{ v\}) - F(C_1)|$, which is large since $C_1$ is a serial circuit and so removing $v$ will destroy the behavior.

\begin{figure}[t]
    \centering
    \includegraphics[width=0.9\textwidth]{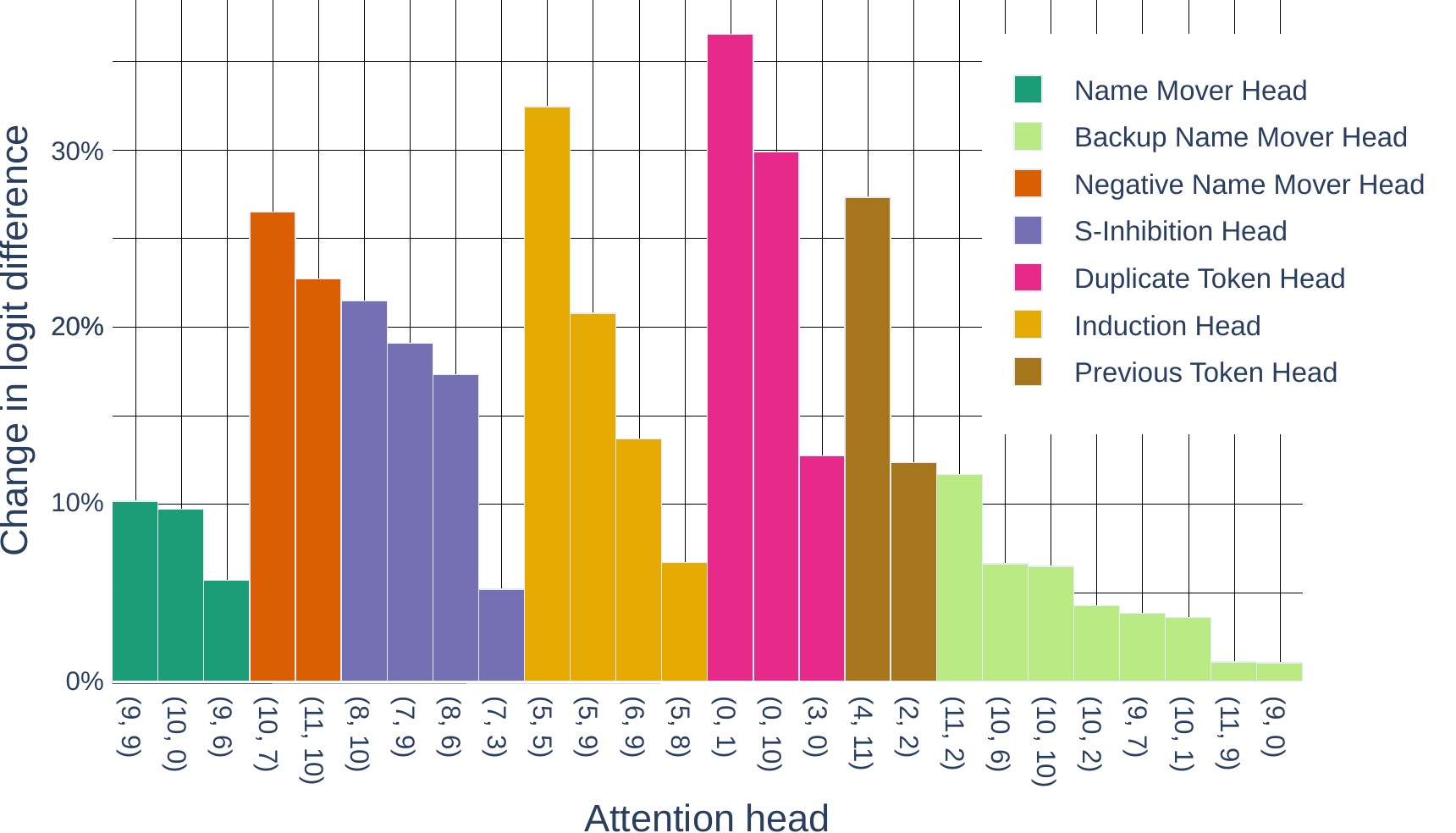}
    \caption{Plot of minimality scores $|F(C \setminus (K \cup \{v\})) - F(C\setminus K)|$ (as percentages of $F(M)$) for all components $v$ in our circuit. The sets $K$ used for each component, as well as the initial and final values of the logit difference for each of these $v$ are in Appendix \ref{app:minimality}. 
    }
    \label{fig:naive_minimality}
\end{figure}

What happens in practice for our circuit? We need to exhibit for every $v$ a set $K$ such that the minimality score is high. 
For most heads, removing the class of heads $G$ that $v$ is a part of provides a reasonable minimality score, but in some instances a more careful choice is needed; we provide full details in Appendix \ref{app:minimality} and display final results in Figure~\ref{fig:naive_minimality}. The importance of different nodes varies, but all have a nontrivial impact (at least 1\% of the original logit difference). These results ensure that we did not interpret irrelevant nodes, but do show that the individual contribution of some attention heads is small. 



\subsection{Comparison with a baseline circuit}
\label{sec:circuit_extr_validation}

In the previous sections, we evaluated our circuit on three quantitative criteria. 
To establish a baseline for these criteria, we compare to a na\"{i}ve circuit that consists of the Name Mover Heads (but no Backup nor Negative Name Mover Heads), S-Inhibition Heads, two Induction Heads, two Duplicate Token Heads, and the Previous Token Heads. This circuit has a faithfulness score of 0.1, comparable to the full circuit $C$. However, the na\"{i}ve circuit can be more easily proven incomplete: by sampling random sets or by knocking-out by classes, we see that $F(M\setminus K)$ is much higher than $F(C\setminus K)$ (Figure \ref{fig:completenessA}). Nonetheless, when we applied the greedy heuristic to optimize for the incompleteness score, both circuits have similarly large incompleteness scores. 

\subsection{Designing adversarial examples}
As argued in \citet{interp_survey}, one way to evaluate the knowledge gained by interpretability work is to use it for downstream applications. In this section, we do this by using knowledge of the circuit to construct simple adversarial examples for the model. 

As presented in Section \ref{sec:legibility}, the model relies on duplicate detection to differentiate between S and IO. 
Motivated by this, we constructed passages where both the S and IO tokens are duplicated. An example is ``John and Mary went to the store. Mary had a good day. John gave a bottle of milk to''; see Appendix~\ref{app:advexes_templates} for full details. We find that this significantly reduces the logit difference and causes the model to predict S over IO $23\%$ of the time (Figure~\ref{table:advexes_results}).

To ensure that the observed effect is not an artifact of the additional sentences, we included a control dataset using the same templates, but where the middle sentence contains S instead of IO. In these sentences, S appears three times in total and IO only appears once. On this distribution, the model has an even higher logit difference than on $\pioi$, and predicts S over IO only $0.4\%$ of the time.




\begin{figure}
\begin{center}
\begin{tabular}{ |c|c|c|c|} 
\hline
\textbf{Distribution} & \textbf{Logit difference} & \textbf{IO probability} &  \textbf{\specialcell{ Proportion of\\S logit greater than IO}}\\ 
\hline 
$\pioi$ & 3.55 & 0.49 & 0.7\%  \\
\hline
\specialcell{Additional occurrence of S\\(natural sentence)} & 3.64 & 0.59 & 0.4\% \\
\hline
\specialcell{Additional occurrence of IO\\(natural sentence)} & 1.23 & 0.36 & 23.4\% \\
\hline
\end{tabular}
\caption{Summary of GPT-2 performance metrics on the IOI task on different datasets. In the line order: for $\pioi$, for the dataset where we added an occurrence of S (thus S appears three times in the sentence) and for the adversarial dataset with duplicated IO in natural sentences.}
\label{table:advexes_results}
\end{center}
\end{figure}

\textbf{Limitations of the attack.}
Despite being inspired by our understanding of our circuit, those examples are simple enough that they could have been found without our circuit with enough effort. 

Moreover, we do not have a full understanding of the mechanisms at play in these adversarial examples. For instance, the S-Inhibition Heads attend not only to S2, but also to the second occurrence of IO. As this pattern is not present in $\pioi$ nor in $\pabc$, it is beyond the analysis presented in Section \ref{sec:legibility}. The study of the behavior of the circuit on these adversarial examples could be a promising area for future work.

\section{Discussion}

In this work, we isolated, understood and validated a set of attention heads in GPT-2 small composed in a circuit that identifies indirect objects. 
Along the way, we discovered interesting structures emerging from the model internals. For instance:
\begin{itemize}
    \item We identified an example of attention heads communicating by using pointers  (sharing the location of a piece of information instead of copying it).
    \item We identified heads compensating for the loss of function of other heads, and heads contributing negatively to the next-token prediction. Early results suggest that the latter phenomenon occurs for other tasks beyond IOI (see Appendix \ref{app:random_tok_seq}).
\end{itemize}


However, our work also has several limitations. First, despite the detailed analysis presented here, we do not understand several components. Those include the attention patterns of the S-Inhibition Heads, and the effect of MLPs and layer norms. 
%
%
Second, the number of parameters in GPT-2 small is orders of magnitude away from state-of-the-art transformer language models. A future challenge is to scale this approach to these larger models.

As a starting point, we performed a preliminary analysis of GPT-2 medium. We found that this model also has a sparse set of heads directly influencing the logits. However, not all of these heads attend to IO and S, suggesting more complex behavior than the Name Movers Heads in GPT-2 small. 
Continuing this investigation is an exciting line of future work. 

More generally, we think that zooming in on a clearly defined task in a single model enables a rich description of phenomena that are likely to be found in broader circumstances. In the same way model organisms in biology enable discoveries valuable beyond a specific species, we think these highly detailed (though narrow) examples are crucial to developing a comprehensive understanding of machine learning models.

\subsubsection*{Acknowledgments}
The authors would like to thank Neel Nanda, Ryan Greenblatt, Lisa Dunlap, Nate Thomas, Stephen Casper, Kayo Yin, Dan Hendrycks, Nora Belrose, Jacob Hilton, Charlie Snell and Lawrence Chan for their useful feedback.


\bibliography{iclr2023_conference}

\begin{thebibliography}{23}
\providecommand{\natexlab}[1]{#1}
\providecommand{\url}[1]{\texttt{#1}}
\expandafter\ifx\csname urlstyle\endcsname\relax
  \providecommand{\doi}[1]{doi: #1}\else
  \providecommand{\doi}{doi: \begingroup \urlstyle{rm}\Url}\fi

\bibitem[Barak et~al.(2022)Barak, Edelman, Goel, Kakade, Malach, and
  Zhang]{barak2022hidden}
Boaz Barak, Benjamin~L Edelman, Surbhi Goel, Sham Kakade, Eran Malach, and
  Cyril Zhang.
\newblock Hidden progress in deep learning: Sgd learns parities near the
  computational limit.
\newblock \emph{arXiv preprint arXiv:2207.08799}, 2022.

\bibitem[Bolukbasi et~al.(2021)Bolukbasi, Pearce, Yuan, Coenen, Reif,
  Vi{\'{e}}gas, and Wattenberg]{interp-illusion-bert}
Tolga Bolukbasi, Adam Pearce, Ann Yuan, Andy Coenen, Emily Reif, Fernanda~B.
  Vi{\'{e}}gas, and Martin Wattenberg.
\newblock An interpretability illusion for {BERT}.
\newblock \emph{CoRR}, abs/2104.07143, 2021.
\newblock URL \url{https://arxiv.org/abs/2104.07143}.

\bibitem[Brown et~al.(2020)Brown, Mann, Ryder, Subbiah, Kaplan, Dhariwal,
  Neelakantan, Shyam, Sastry, Askell, Agarwal, Herbert-Voss, Krueger, Henighan,
  Child, Ramesh, Ziegler, Wu, Winter, Hesse, Chen, Sigler, Litwin, Gray, Chess,
  Clark, Berner, McCandlish, Radford, Sutskever, and Amodei]{GPT3}
Tom Brown, Benjamin Mann, Nick Ryder, Melanie Subbiah, Jared~D Kaplan, Prafulla
  Dhariwal, Arvind Neelakantan, Pranav Shyam, Girish Sastry, Amanda Askell,
  Sandhini Agarwal, Ariel Herbert-Voss, Gretchen Krueger, Tom Henighan, Rewon
  Child, Aditya Ramesh, Daniel Ziegler, Jeffrey Wu, Clemens Winter, Chris
  Hesse, Mark Chen, Eric Sigler, Mateusz Litwin, Scott Gray, Benjamin Chess,
  Jack Clark, Christopher Berner, Sam McCandlish, Alec Radford, Ilya Sutskever,
  and Dario Amodei.
\newblock Language models are few-shot learners.
\newblock In H.~Larochelle, M.~Ranzato, R.~Hadsell, M.F. Balcan, and H.~Lin
  (eds.), \emph{Advances in Neural Information Processing Systems}, volume~33,
  pp.\  1877--1901. Curran Associates, Inc., 2020.
\newblock URL
  \url{https://proceedings.neurips.cc/paper/2020/file/1457c0d6bfcb4967418bfb8ac142f64a-Paper.pdf}.

\bibitem[Caldarini et~al.(2022)Caldarini, Jaf, and
  McGarry]{caldarini2022literature}
Guendalina Caldarini, Sardar Jaf, and Kenneth McGarry.
\newblock A literature survey of recent advances in chatbots.
\newblock \emph{Information}, 13\penalty0 (1):\penalty0 41, 2022.

\bibitem[Elhage et~al.(2021)Elhage, Nanda, Olsson, Henighan, Joseph, Mann,
  Askell, Bai, Chen, Conerly, DasSarma, Drain, Ganguli, Hatfield-Dodds,
  Hernandez, Jones, Kernion, Lovitt, Ndousse, Amodei, Brown, Clark, Kaplan,
  McCandlish, and Olah]{elhage2021mathematical}
Nelson Elhage, Neel Nanda, Catherine Olsson, Tom Henighan, Nicholas Joseph, Ben
  Mann, Amanda Askell, Yuntao Bai, Anna Chen, Tom Conerly, Nova DasSarma, Dawn
  Drain, Deep Ganguli, Zac Hatfield-Dodds, Danny Hernandez, Andy Jones, Jackson
  Kernion, Liane Lovitt, Kamal Ndousse, Dario Amodei, Tom Brown, Jack Clark,
  Jared Kaplan, Sam McCandlish, and Chris Olah.
\newblock A mathematical framework for transformer circuits.
\newblock \emph{Transformer Circuits Thread}, 2021.
\newblock https://transformer-circuits.pub/2021/framework/index.html.

\bibitem[Geiger et~al.(2021)Geiger, Lu, Icard, and Potts]{geiger2021causal}
Atticus Geiger, Hanson Lu, Thomas~F Icard, and Christopher Potts.
\newblock Causal abstractions of neural networks.
\newblock In A.~Beygelzimer, Y.~Dauphin, P.~Liang, and J.~Wortman Vaughan
  (eds.), \emph{Advances in Neural Information Processing Systems}, 2021.
\newblock URL \url{https://openreview.net/forum?id=RmuXDtjDhG}.

\bibitem[Geva et~al.(2020)Geva, Schuster, Berant, and
  Levy]{geva2020transformer}
Mor Geva, Roei Schuster, Jonathan Berant, and Omer Levy.
\newblock Transformer feed-forward layers are key-value memories.
\newblock \emph{arXiv preprint arXiv:2012.14913}, 2020.

\bibitem[Hendrycks \& Mazeika(2022)Hendrycks and Mazeika]{Hendrycks2022XRiskAF}
Dan Hendrycks and Mantas Mazeika.
\newblock X-risk analysis for ai research.
\newblock \emph{arXiv}, abs/2206.05862, 2022.

\bibitem[Hernandez et~al.(2021)Hernandez, Schwettmann, Bau, Bagashvili,
  Torralba, and Andreas]{hernandez2021natural}
Evan Hernandez, Sarah Schwettmann, David Bau, Teona Bagashvili, Antonio
  Torralba, and Jacob Andreas.
\newblock Natural language descriptions of deep visual features.
\newblock In \emph{International Conference on Learning Representations}, 2021.

\bibitem[Jain \& Wallace(2019)Jain and Wallace]{attention_not_explanation}
Sarthak Jain and Byron~C. Wallace.
\newblock {A}ttention is not {E}xplanation.
\newblock In \emph{Proceedings of the 2019 Conference of the North {A}merican
  Chapter of the Association for Computational Linguistics: Human Language
  Technologies, Volume 1 (Long and Short Papers)}, pp.\  3543--3556,
  Minneapolis, Minnesota, June 2019. Association for Computational Linguistics.
\newblock \doi{10.18653/v1/N19-1357}.
\newblock URL \url{https://aclanthology.org/N19-1357}.

\bibitem[Meng et~al.(2022)Meng, Bau, Andonian, and Belinkov]{meng2022locating}
Kevin Meng, David Bau, Alex Andonian, and Yonatan Belinkov.
\newblock Locating and editing factual associations in gpt.
\newblock \emph{arXiv preprint arXiv:2202.05262}, 2022.

\bibitem[Michel et~al.(2019)Michel, Levy, and Neubig]{michel2019sixteen}
Paul Michel, Omer Levy, and Graham Neubig.
\newblock Are sixteen heads really better than one?
\newblock \emph{Advances in neural information processing systems}, 32, 2019.

\bibitem[Mu \& Andreas(2020)Mu and Andreas]{mu2020compositionalExplanations}
Jesse Mu and Jacob Andreas.
\newblock Compositional explanations of neurons.
\newblock \emph{Advances in Neural Information Processing Systems},
  33:\penalty0 17153--17163, 2020.

\bibitem[Nanda \& Lieberum(2022)Nanda and Lieberum]{meca_interp_grokking}
Neel Nanda and Tom Lieberum.
\newblock A mechanistic interpretability analysis of grokking, 2022.
\newblock URL
  \url{https://www.alignmentforum.org/posts/N6WM6hs7RQMKDhYjB/a-mechanistic-interpretability-analysis-of-grokking}.

\bibitem[Olah(2022)]{AnthropicMechanisticEssay}
Chris Olah.
\newblock Mechanistic interpretability, variables, and the importance of
  interpretable bases.
\newblock \url{https://www.transformer-circuits.pub/2022/mech-interp-essay},
  2022.
\newblock Accessed: 2022-15-09.

\bibitem[Olah et~al.(2020)Olah, Cammarata, Schubert, Goh, Petrov, and
  Carter]{olah2020zoom}
Chris Olah, Nick Cammarata, Ludwig Schubert, Gabriel Goh, Michael Petrov, and
  Shan Carter.
\newblock Zoom in: An introduction to circuits.
\newblock \emph{Distill}, 2020.
\newblock \doi{10.23915/distill.00024.001}.
\newblock https://distill.pub/2020/circuits/zoom-in.

\bibitem[Olsson et~al.(2022)Olsson, Elhage, Nanda, Joseph, DasSarma, Henighan,
  Mann, Askell, Bai, Chen, et~al.]{olsson2022context}
Catherine Olsson, Nelson Elhage, Neel Nanda, Nicholas Joseph, Nova DasSarma,
  Tom Henighan, Ben Mann, Amanda Askell, Yuntao Bai, Anna Chen, et~al.
\newblock In-context learning and induction heads.
\newblock \emph{arXiv preprint arXiv:2209.11895}, 2022.

\bibitem[Radford et~al.(2019)Radford, Wu, Child, Luan, Amodei, and
  Sutskever]{GPT2}
Alec Radford, Jeff Wu, Rewon Child, David Luan, Dario Amodei, and Ilya
  Sutskever.
\newblock Language models are unsupervised multitask learners.
\newblock 2019.

\bibitem[Räuker et~al.(2022)Räuker, Ho, Casper, and
  Hadfield-Menell]{interp_survey}
Tilman Räuker, Anson Ho, Stephen Casper, and Dylan Hadfield-Menell.
\newblock Toward transparent ai: A survey on interpreting the inner structures
  of deep neural networks, 2022.
\newblock URL \url{https://arxiv.org/abs/2207.13243}.

\bibitem[Vaswani et~al.(2017)Vaswani, Shazeer, Parmar, Uszkoreit, Jones, Gomez,
  Kaiser, and Polosukhin]{Attention}
Ashish Vaswani, Noam Shazeer, Niki Parmar, Jakob Uszkoreit, Llion Jones,
  Aidan~N Gomez, {\L}ukasz Kaiser, and Illia Polosukhin.
\newblock Attention is all you need.
\newblock In \emph{Advances in Neural Information Processing Systems}, pp.\
  5998--6008, 2017.

\bibitem[Vig et~al.(2020)Vig, Gehrmann, Belinkov, Qian, Nevo, Singer, and
  Shieber]{vig2020investigating}
Jesse Vig, Sebastian Gehrmann, Yonatan Belinkov, Sharon Qian, Daniel Nevo,
  Yaron Singer, and Stuart Shieber.
\newblock Investigating gender bias in language models using causal mediation
  analysis.
\newblock \emph{Advances in Neural Information Processing Systems},
  33:\penalty0 12388--12401, 2020.

\bibitem[Wei et~al.(2022)Wei, Tay, Bommasani, Raffel, Zoph, Borgeaud, Yogatama,
  Bosma, Zhou, Metzler, Chi, Hashimoto, Vinyals, Liang, Dean, and
  Fedus]{Wei2022EmergentAO}
Jason Wei, Yi~Tay, Rishi Bommasani, Colin Raffel, Barret Zoph, Sebastian
  Borgeaud, Dani Yogatama, Maarten Bosma, Denny Zhou, Donald Metzler, Ed~Chi,
  Tatsunori Hashimoto, Oriol Vinyals, Percy Liang, Jeff Dean, and William
  Fedus.
\newblock Emergent abilities of large language models.
\newblock \emph{ArXiv}, abs/2206.07682, 2022.

\bibitem[Zhang et~al.(2022)Zhang, Xing, Zou, and Wu]{zhang2022shifting}
Angela Zhang, Lei Xing, James Zou, and Joseph~C Wu.
\newblock Shifting machine learning for healthcare from development to
  deployment and from models to data.
\newblock \emph{Nature Biomedical Engineering}, pp.\  1--16, 2022.

\end{thebibliography}
\bibliographystyle{iclr2023_conference}

\section{Appendix}

\appendix

\section{Disentangling token and positional signal in the output of S-Inhibition Heads} \label{app:disantangling_signals}

In Section \ref{sec:s_inhibition}, we discovered that S-Inhibition Heads are responsible for the Name Mover Heads' specific attention on the IO token, and in particular we discovered that their output is a signal for Name Movers Heads to avoid the S1 token. 

More precisely, we discover that they were outputting \emph{token signals} (information about the value of the token S) and \emph{positional signals} (related to the value of the position S1). We describe here the details of this discovery.

To disentangle the two effects, we design a series of counterfactual datasets where only some signals are present, and some are inverted with respect to the original dataset. In this section, we conducted activation patching (or simply patching). Instead of patching isolated paths, we patch the output of a set of heads and recompute the forward pass of the model after this modification. As shown in Section \ref{sec:distributed_name_movers}, the output of later (Backup) Name Mover Heads and Negative Heads depend on earlier heads from these classes, such that it is not possible to ignore the interactions between heads inside a class (as it is the case in path patching in Appendix \ref{app:path_patching}).

By patching the output of S-Inhibition heads from these datasets, we can quantify in isolation the impact of each signal on the final logit difference.

We constructed six datasets by combining three transformations of the original $\pioi$ distribution.
\begin{compactitem}
    \item \textbf{Random name flip}: we replace the names from a given sentence with random names, but we keep the same position for all names. Moreover, each occurrence of a name in the original sentence is replaced by the same random name. When we patch outputs of S-Inhibition heads from this sentence, only positional signals are present, the token signals are unrelated to the names of the original sequence.
    \item \textbf{IO$\leftrightarrow$S1 flip}: we swap the position of IO and S1. The output of S-inhibition heads will contain correct token signals (the subject of the second clause is the same) but inverted positional signals (because the position of IO and S1 are swapped)
    \item \textbf{IO$\leftarrow$S2 replacement}: we make IO become the subject of the sentence and S the indirect object. In this dataset, both token signals and positional signals are inverted.
\end{compactitem}
We can also compose these transformations. For instance, we can create a dataset with no token signals and inverted positional signals by applying IO$\leftrightarrow$S1 flip on the dataset with random names. In total, we can create all six combinations of original, inverted, or uncorrelated token signal with the original and inverted positional signal.

From each of those six datasets, we patched the output of S-Inhibition heads and measured the logit difference. The results are presented in Figure \ref{table:signals_logit_diff}.

\begin{figure}
\begin{center}
\begin{tabular}{ |c|c|c|c|} 
\hline
\textbf{~} & \textbf{Original positional signal} & \textbf{Inverted position signal} \\ 
\hline 
Original S token signal & 3.55 (baseline) & -0.99  \\
\hline
Random S token signal& 2.45 & -1.96  \\
\hline
S$\leftrightarrow$IO inverted token signal & 1.77 & -3.16  \\
\hline
\end{tabular}
\caption{Logit difference after patching S-Inhibition heads from signal-specific datasets. The effect on logit difference can be decomposed as a sum of the effects of position and token signal. }
\label{table:signals_logit_diff}
\end{center}
\end{figure}

\begin{figure}
    \centering

     \includegraphics[width=0.7\textwidth]{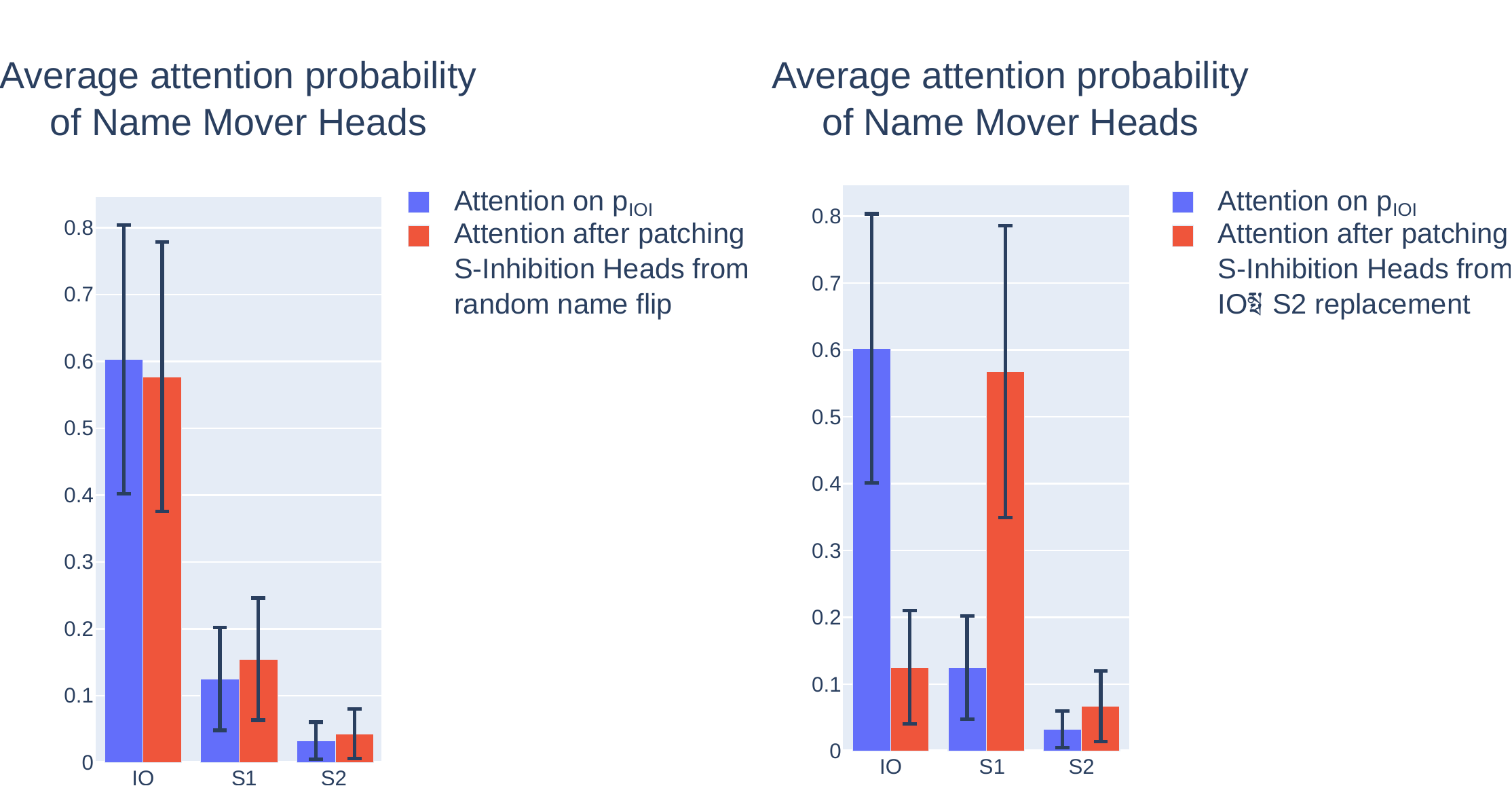}

    \caption{Name Mover Heads' attention probability before and after patching S-Inhibition Heads from signal-specific datasets. Left: patching from the dataset generated by random flip of name (same position signal, random token signal). Right: patching from the dataset generated by IO$\leftarrow$S2 replacement (inverted position signal, inverted token signal). Black bars represent the standard deviation.   }
    \label{fig:effect_s_in_patching_nm_attention}
\end{figure}

These results can be summarized as the sum of the two effects. Suppose we define the variable $S_{tok}$ to be 1 if the token signal is the original, 0 when uncorrelated and -1 when inverted. And similarly $S_{pos}$ to be 1 if the position signal is the original and -1 if inverted. Then the Figure \ref{table:signals_logit_diff} suggests that the logit difference can be well approximated by $2.31S_{pos} + 0.99S_{tok}$, with a mean error of 7\% relative to the baseline logit difference.

For instance, when both the positional and token signals are inverted, the logit difference is the opposite of the baseline. This means that the S token is predicted stronger than the IO token, as strong as IO before patching. In this situation, due to the contradictory information contained in the output of S-Inhibition heads, the Name Movers attend and copy the S1 token instead of the IO token (see Figure \ref{fig:effect_s_in_patching_nm_attention}, right). 
In the intermediate cases where only one of the signals is modified, we observe a partial effect compared to the fully inverted case (e.g. Figure \ref{fig:effect_s_in_patching_nm_attention}, left). The effect size depends on the altered signals: positional signals are more important than token signals. 

Can we be more specific as to what the token and positional signals are? Unfortunately, we do not have a complete answer, but see this as one of the most interesting further directions of our work. We expect that the majority of the positional information is about the \emph{relative} positional embedding between S1 and S2 (such pointer arithmetic behavior has already been observed in \citet{olsson2022context}). When patching in S2 Inhibition outputs from a distribution where prefixes to sentences are longer (but the distance between S1 and S2 is constant), the logit difference doesn't change (3.56 before patching vs 3.57 after).  This suggests that the positional signal doesn't depend on the absolute position of the tokens, as long as the relative position of S1 and S2 stays the same.


\section{Path patching}
\label{app:path_patching}





\begin{figure}
    \centering
    \includegraphics[width=\textwidth]{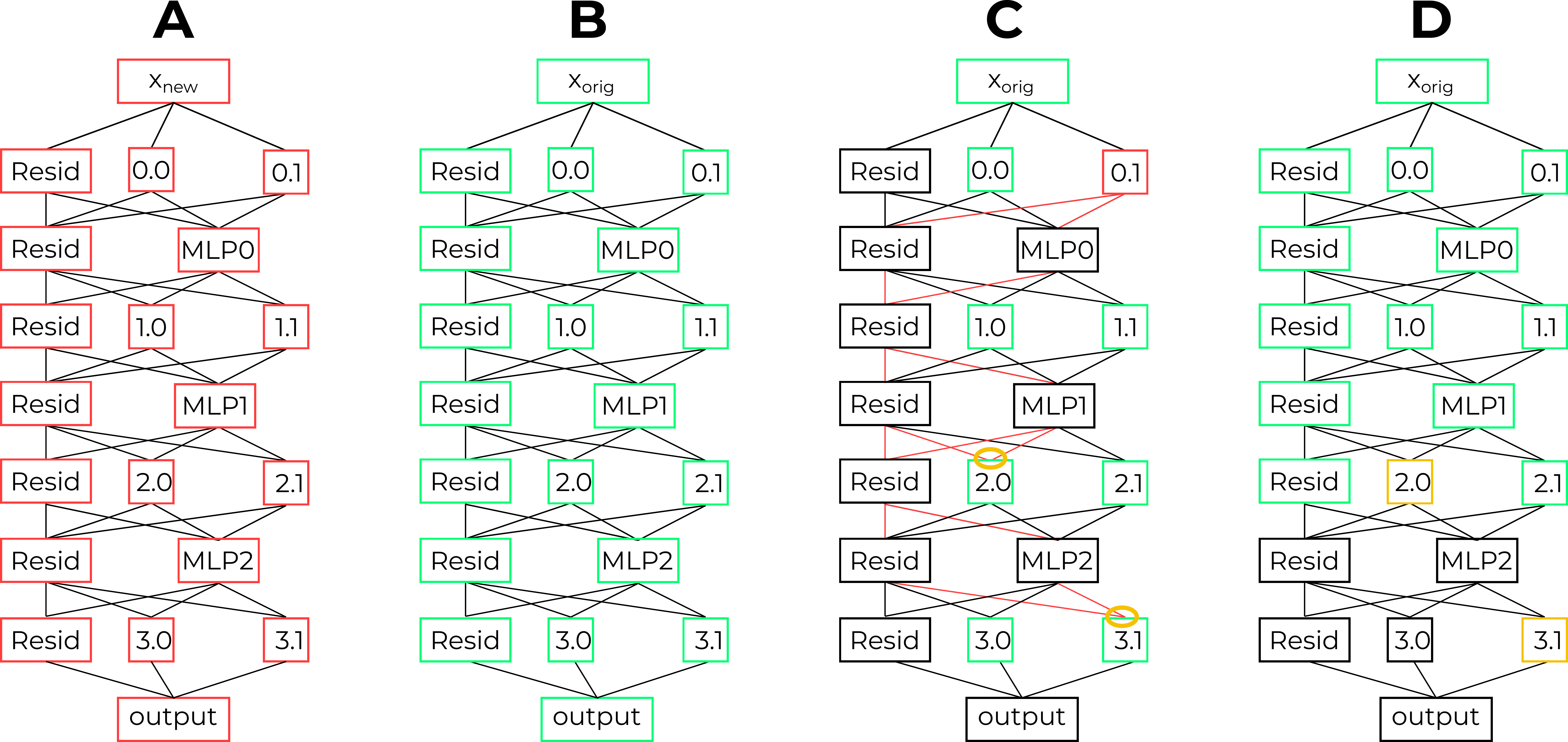}
    \caption{Illustration of the four forward passes in the path patching method. $h=0.1$ and $R=\{2.0, 3.1\}$. The layer norms are not shown for clarity. In the forward pass C, we show in red all the paths through which $h$ can influence the value of heads in $R$. Nodes in black are recomputed, nodes in color are patched or frozen to their corresponding values.}
    \label{fig:path_patching}
\end{figure}

In this section we describe the technical details of the path patching technique we used in many experiments. This technique depends on

\begin{itemize}
    \item $x_{\text{orig}}$ the original data point,
    \item $x_{\text{new}}$ the new data point,
    \item $h$ the \textit{sender} attention head, and
    \item $R \subseteq M$ the set of \textit{receiver} nodes in the model's computational graph $M$. In our case, $R$ is either the input (key, query or values) of a set of attention heads or the end state of the residual stream\footnote{The model uses the end state of the residual stream to compute the logits (after a layer norm), so we choose $R$ equal to this end state when we wish to measure the direct effect of $h$ on the logits.}.
\end{itemize}

It is then implemented as Algorithm \ref{path_patching}, which can be summarized in five steps.

\begin{enumerate}
    \item Gather activations on $x_{\text{orig}}$ and $x_{\text{new}}$. [Lines 1-2, Figure \ref{fig:path_patching}A-B]
    \item Freeze\footnote{We use `freezing' of an activation $a$ to refer to the special case of patching when we begin a forward pass on an input $x$, edit some earlier activations, and then want to patch $a$ to its value on a forward pass of $x$ (which did not have the edits of earlier activations).} all the heads to their activation on $x_{\text{orig}}$ except $h$ that is patched to its activation on $x_{\text{new}}$. [Lines 4-5, Figure \ref{fig:path_patching}C]
    \item Run a forward pass of the model on $x_{\text{orig}}$ with the frozen and patched nodes (MLPs and layer norm are recomputed). [Line 7, Figure \ref{fig:path_patching}C]
    \item In this forward pass, save the activation of the model components $r \in R$ as if they were recomputed (but this value is overwritten in the forward pass because they are frozen). [Line 6-9, Figure \ref{fig:path_patching}C]
    \item Run a last forward pass on $x_{\text{orig}}$ patching the receiver nodes in $R$ to the saved values. [Lines 12-20,  Figure \ref{fig:path_patching}D]
\end{enumerate}

Each of the $A$ variables in Algorithm \ref{path_patching} ($A_{\mathrm{orig}}$, $A_{\mathrm{new}}$ and $A_{\mathrm{patch}}$) stores activations for all nodes in $M$. $A_{\mathrm{orig}}$ and $A_{\mathrm{new}}$ contain output activations while $A_{\mathrm{patch}}$ contains the activation of the nodes in $R$ (keys, queries or values of attention heads, or the end state of the residual stream).

In forward pass C of the algorithm all attention heads except $h$ have their output frozen to their value on $x_{\text{orig}}$. This means that we remove the influence of every path including one (or more) intermediate attention heads $p$ of the form $h \to p \to r$, where $r \in R$. In all cases in the paper, we measure the logit difference on the forward pass D compared to the logit difference of the model, as a measure of the importance of the path $h \to r$. We additionally compute this logit difference as an average over $N>200$ pairs $(\xorig, \xnew)$.


\begin{algorithm}
\caption{Path patching}\label{path_patching}
\begin{algorithmic}[1]
\State Compute all activations $A_{\mathrm{new}}$ on $x_{\text{new}}$ (forward pass A)
\State Compute all activations $A_{\mathrm{orig}}$ on $x_{\text{orig}}$ (forward pass B)
\For{$r \in R$}
    \State $A^r_{\text{patch}} \gets A_{\mathrm{orig}}$
    \State $A^r_{\text{patch}}[h] \gets A^r_{\text{new}}[h]$
    \For{each MLP $m$ between $h$ and $r$}
        \State Recompute $A^r_{\text{patch}}[m]$ from the activations in $A^r_{\text{patch}}$ (forward pass C) 
    \EndFor
    \State Recompute $A^r_{\text{patch}}[r]$ from the activations in $A^r_{\text{patch}}$ (forward pass C)
\EndFor\\

\State $A_{\text{final}} \gets \emptyset$
\For{$c \in M$ nodes of the computational graph, in topologically sorted order,} (forward pass D)
    \If{$c \in R$}
        \State $A_{\text{final}}[c] \gets A^c_{\text{patch}}[c] $
    
    \Else
        \State Compute $A_{\text{final}}[c]$ from activations in $A_{\text{final}}$ 
    \EndIf
\EndFor \\

\Return $A_{\text{final}}[\text{Logits}]$ 
\end{algorithmic}
\end{algorithm}

\section{Direct effect on S-Inhibition Heads' keys} \label{app:s_in_keys}

In this section, we present the direct effect analysis of the S-Inhibition Heads' keys. The experiment is similar to the investigation of the S-Inhibition Heads' values presented in Section \ref{sec:s_inhibition}.

\begin{figure}
    \centering
     \begin{subfigure}[b]{0.43\textwidth}
         \centering
         \includegraphics[width=\textwidth]{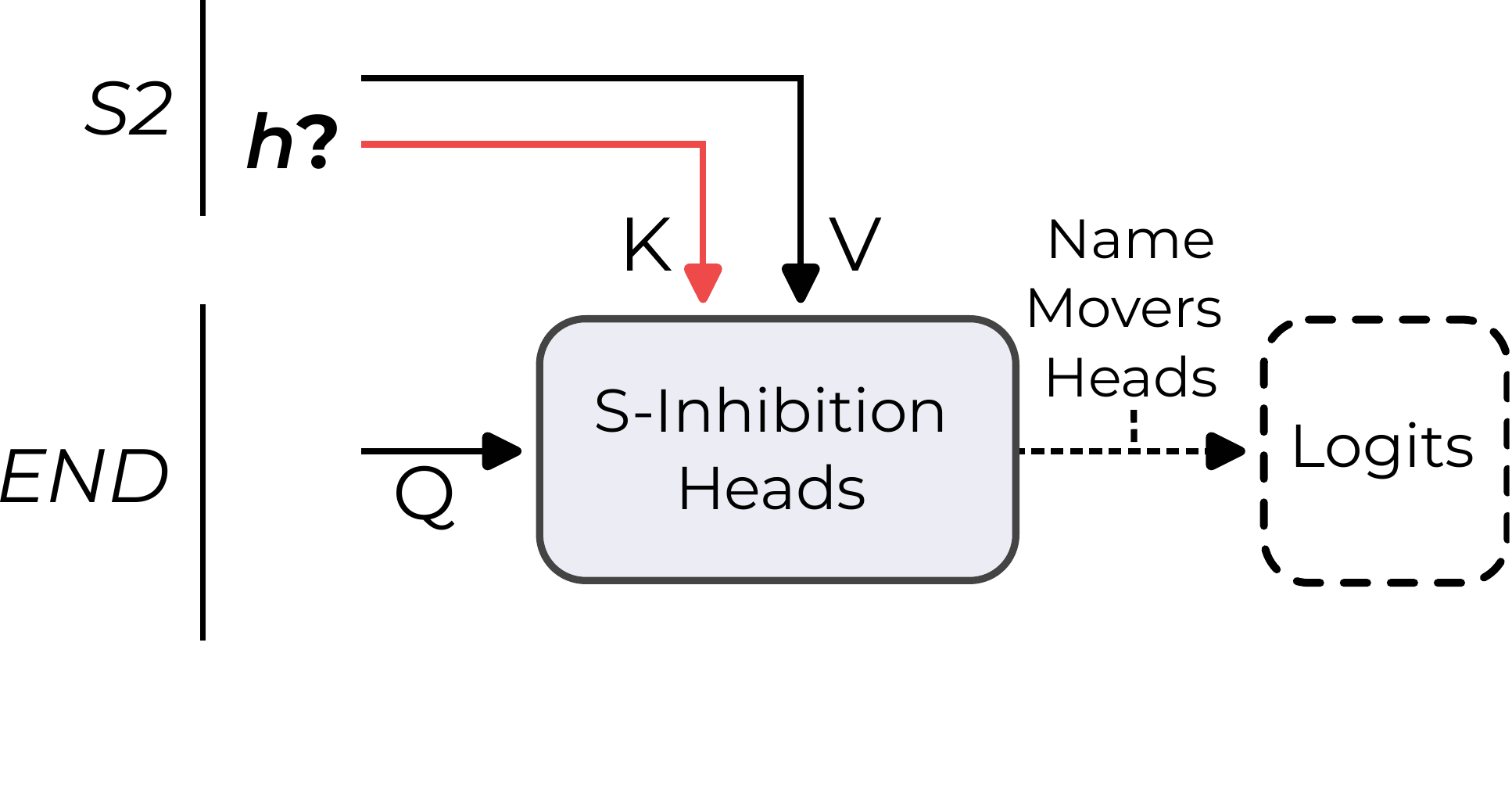}
         \caption{}
         \label{fig:effect_on_sin_keysA}
     \end{subfigure}
     \hfill
     \begin{subfigure}[b]{0.43\textwidth}
         \centering
         \includegraphics[width=\textwidth]{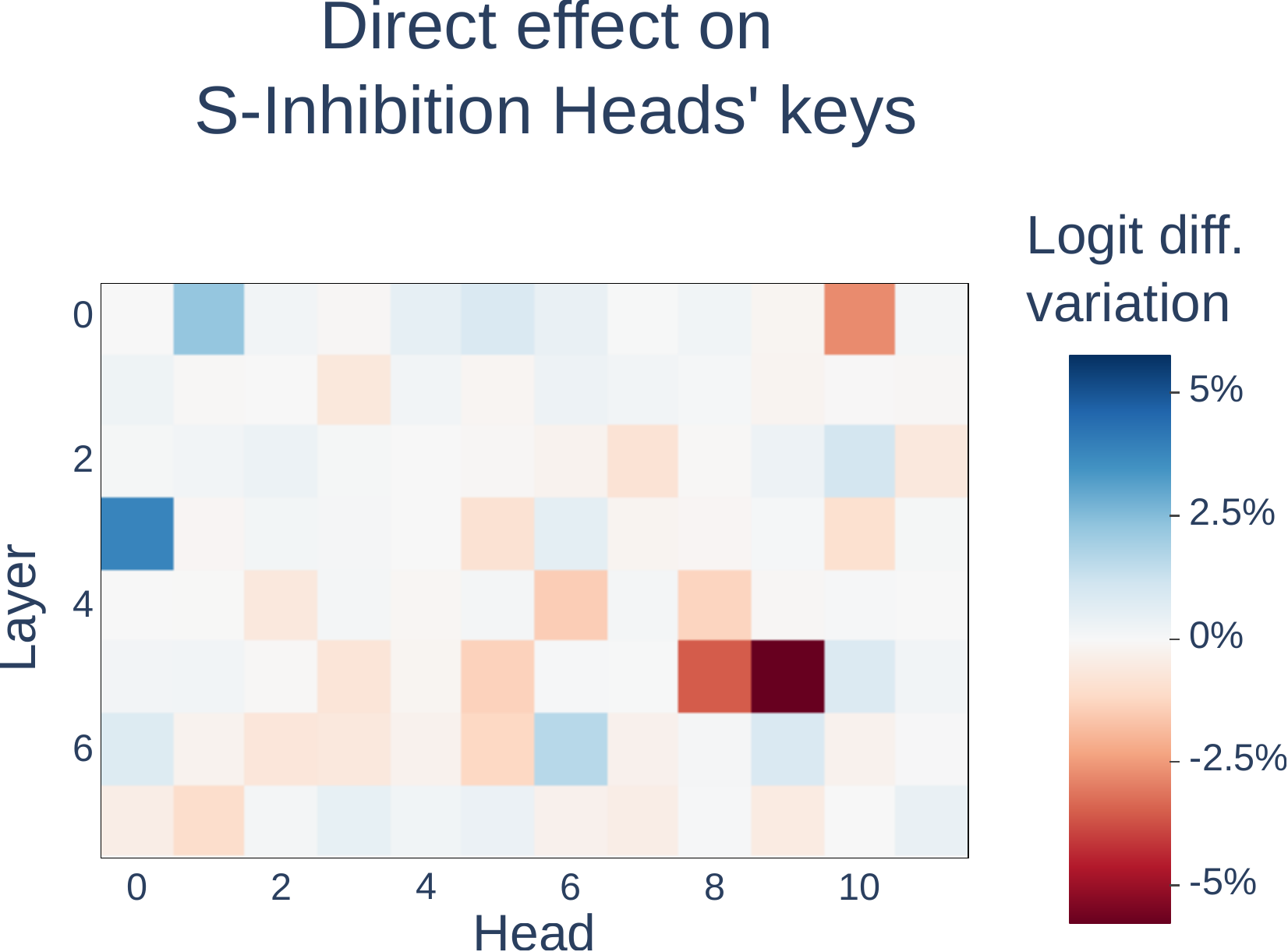}
         \caption{}
         \label{fig:effect_on_sin_keysB}
     \end{subfigure}
    \caption{(a) Diagram of the direct effect experiment on S-Inhibition Heads' keys at S2. (b) Result of the path patching experiments for direct effect experiment on S-Inhibition Heads' keys.}
    \label{fig:effect_on_sin_keys}
\end{figure}

The results presented in Figure \ref{fig:effect_on_sin_keysB} show that some heads significantly influence the logit difference through S-Inhibition Heads' keys. We observe that Duplicate Token Heads (3.0 and 0.1) appear to also influence S-Inhibition Heads' values, but their effect is reversed.

Moreover, we identify 3 new heads influencing positively the logit difference: 5.9, 5.8 and 0.10. 

\textbf{Fuzzy Duplicate Token Head} By looking at its attention pattern, we identified that 0.10 was paying attention to S1 from S2. However, the attention pattern was fuzzy, as intermediate tokens also have non-negligible attention probability. We thus call it a fuzzy Duplicate Token Head. On Open Web Text (OWT), the fuzzy Duplicate Token Head attend to duplicates in a short range (see Appendix~\ref{app:validation_dups}).

\textbf{Fuzzy Induction Heads} The heads 5.9, 5.8 are paying attention to S1+1. For 5.8, S1+1 is the token with the highest attention probability after the start token. But the absolute value is small (less than 0.1). The head 5.9 is paying attention to S1+1 but also to tokens before S. Because of these less interpretable attention patterns, we called them fuzzy Induction Heads. (see Appendix~\ref{app:random_tok_seq} for more details about their behavior outside IOI).

By influencing the keys of the S-Inhibition Heads, we hypothesize that those heads are amplifying the positional signal written by the other Induction and Duplicate Token Heads.

\section{Identification of Previous Token Heads} \label{app:prev_token_heads}

\begin{figure}
    \centering
     \begin{subfigure}[b]{0.43\textwidth}
         \centering
         \includegraphics[width=\textwidth]{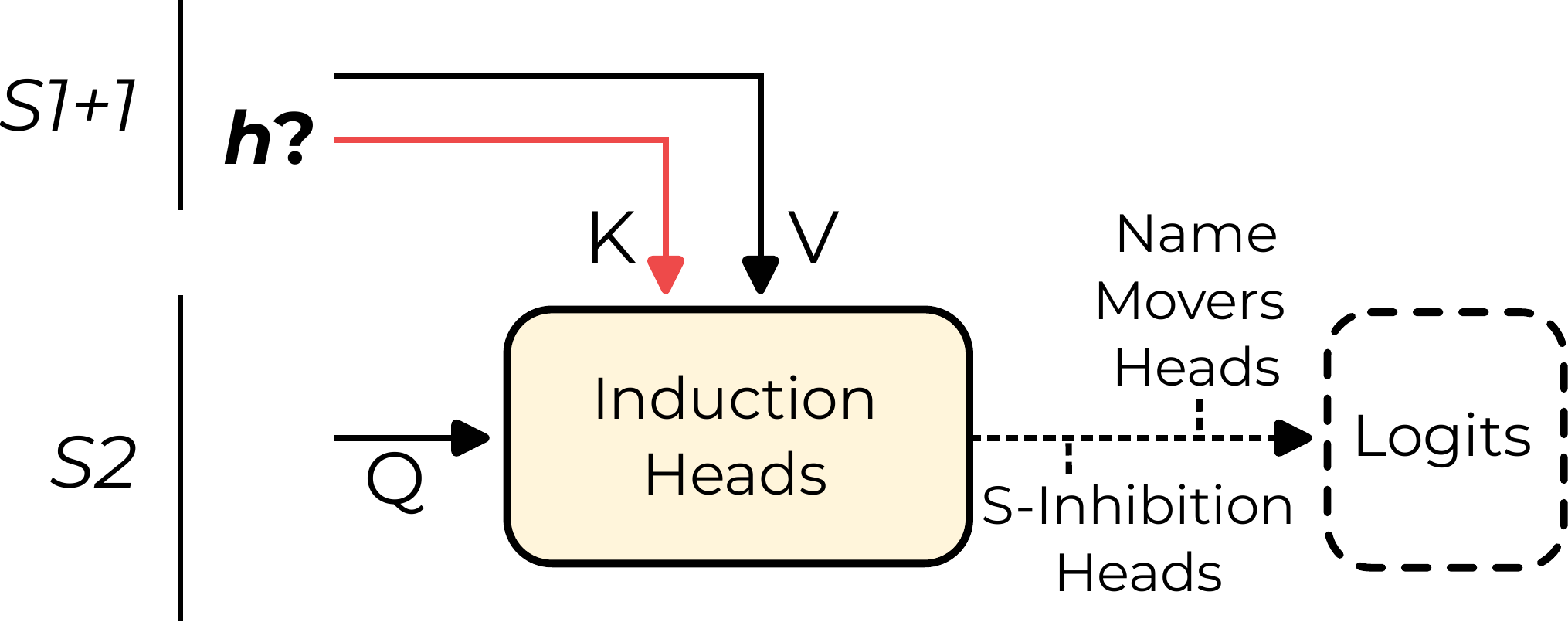}
         \caption{}
         \label{fig:prev_tokA}
     \end{subfigure}
     \hfill
     \begin{subfigure}[b]{0.43\textwidth}
         \centering
         \includegraphics[width=\textwidth]{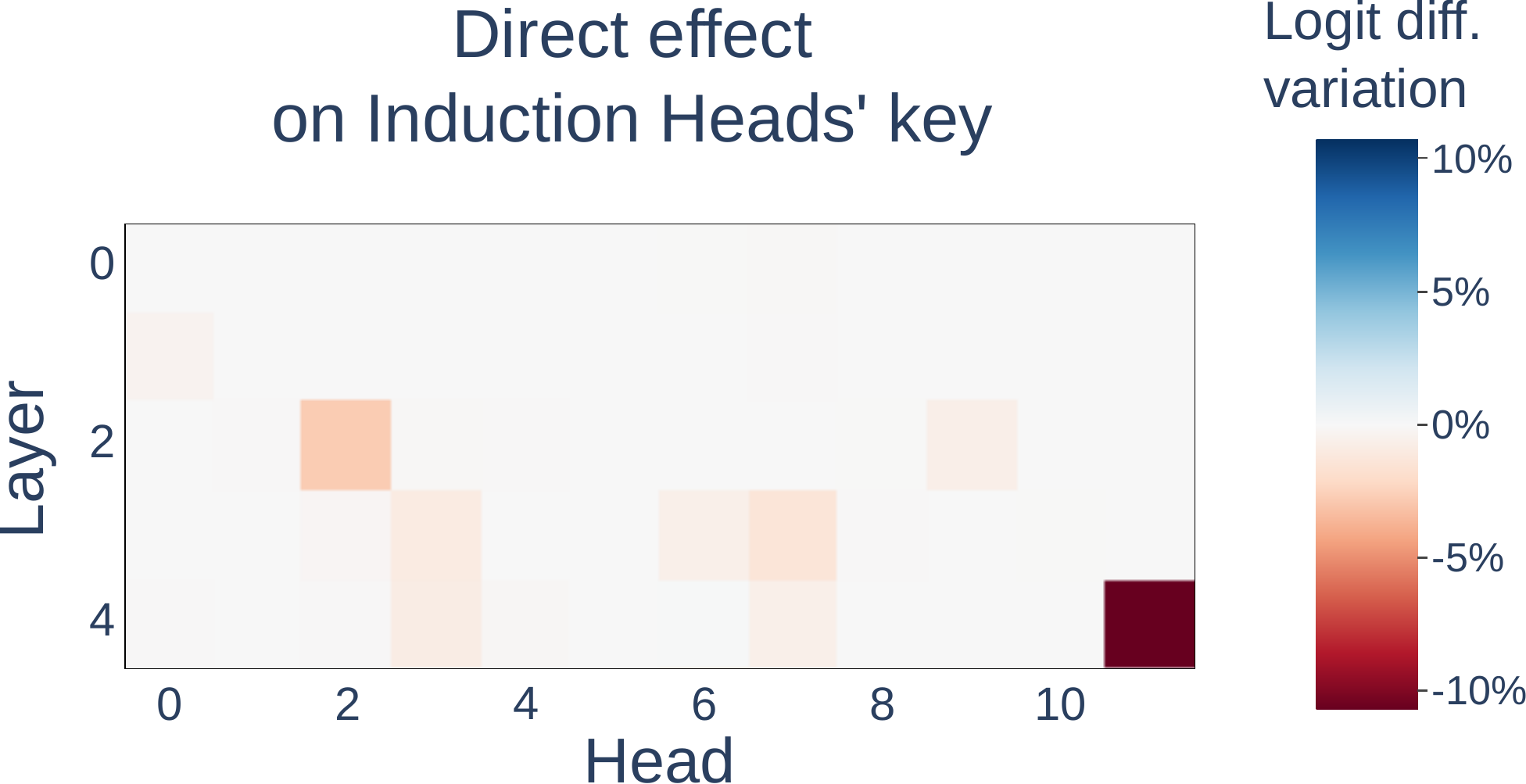}
         \caption{}
         \label{fig:prev_tokB}
     \end{subfigure}
    \caption{(a) Diagram of the direct effect experiment on Induction Heads' keys at S1+1. The effects on logits are mediated by S-Inhibition Heads and Name Mover Heads (b) Result of the path patching experiments on Induction Heads' keys at S1+1.}
    \label{fig:prev_tok}
\end{figure}

Induction Heads rely on key composition with Previous Token Heads to recognize patterns of the form \texttt{[A]\,[B]\,...\,[A]}. In the context of IOI, the repeated token \texttt{[A]} is S2. We thus searched for heads directly affecting Induction Heads keys at the S1+1 position (the \texttt{[B]} token in the general pattern). For this, we used path patching. The results of the experiment are visible in Figure \ref{fig:prev_tok}. We identify two main heads causing a decrease in logit difference (and thus contributing positively to the logit difference): 4.11 and 2.2. These heads pay primary attention to the previous token. This is coherent with the Previous Token Heads we were expecting. We thoroughly investigate their attention patterns outside of $\pioi$ in Appendix \ref{app:random_tok_seq}.

\citet{olsson2022context} also describes an induction mechanism relying on query composition in GPT-2. We performed path patching to investigate heads influencing the Induction Heads queries at the S2 position, but did not find any significant effect.

\section{IOI Templates} 
\label{app:templates}
We list all the templates we used in Figure \ref{fig:table_templates}. Each name was drawn from a list of 100 English first names, while the place and the object were chosen among a hand-made list of 20 common words. All the words chosen were one token long to ensure proper sequence alignment computation of the mean activations.

\begin{figure}[H]
\centering
\begin{tabular}{ |c| } 
\hline
Templates in $\pioi$ \\ 
\hline 
Then, [B] and [A] went to the [PLACE]. [B] gave a [OBJECT] to [A] \\
\hline
Then, [B] and [A] had a lot of fun at the [PLACE]. [B] gave a [OBJECT] to [A] \\
\hline
Then, [B] and [A] were working at the [PLACE]. [B] decided to give a [OBJECT] to [A] \\
\hline
Then, [B] and [A] were thinking about going to the [PLACE]. [B] wanted to give a [OBJECT] to [A] \\
\hline
Then, [B] and [A] had a long argument, and afterwards [B] said to [A] \\
\hline

After [B] and [A] went to the [PLACE], [B] gave a [OBJECT] to [A] \\
\hline

When [B] and [A] got a [OBJECT] at the [PLACE], [B] decided to give it to [A] \\
\hline

When [B] and [A] got a [OBJECT] at the [PLACE], [B] decided to give the [OBJECT] to [A] \\
\hline

While [B] and [A] were working at the [PLACE], [B] gave a [OBJECT] to [A] \\
\hline
While [B] and [A] were commuting to the [PLACE], [B] gave a [OBJECT] to [A] \\
\hline
After the lunch, [B] and [A] went to the [PLACE]. [B] gave a [OBJECT] to [A] \\
\hline
Afterwards, [B] and [A] went to the [PLACE]. [B] gave a [OBJECT] to [A] \\
\hline

Then, [B] and [A] had a long argument. Afterwards [B] said to [A] \\
\hline

The [PLACE] [B] and [A] went to had a [OBJECT]. [B] gave it to [A] \\
\hline

Friends [B] and [A] found a [OBJECT] at the [PLACE]. [B] gave it to [A] \\
\hline
\end{tabular}
\caption{Templates used in the IOI dataset. All templates in the table fit the `BABA' pattern, but we use templates that fit the `ABBA' pattern as well (i.e, by swapping the first instances of [B] and [A] in all of the above).}
\label{fig:table_templates}
\end{figure}

\section{Backup Name Mover Heads} \label{app:backup_name_movers}

\begin{figure}
    \centering
    \includegraphics[width=\textwidth]{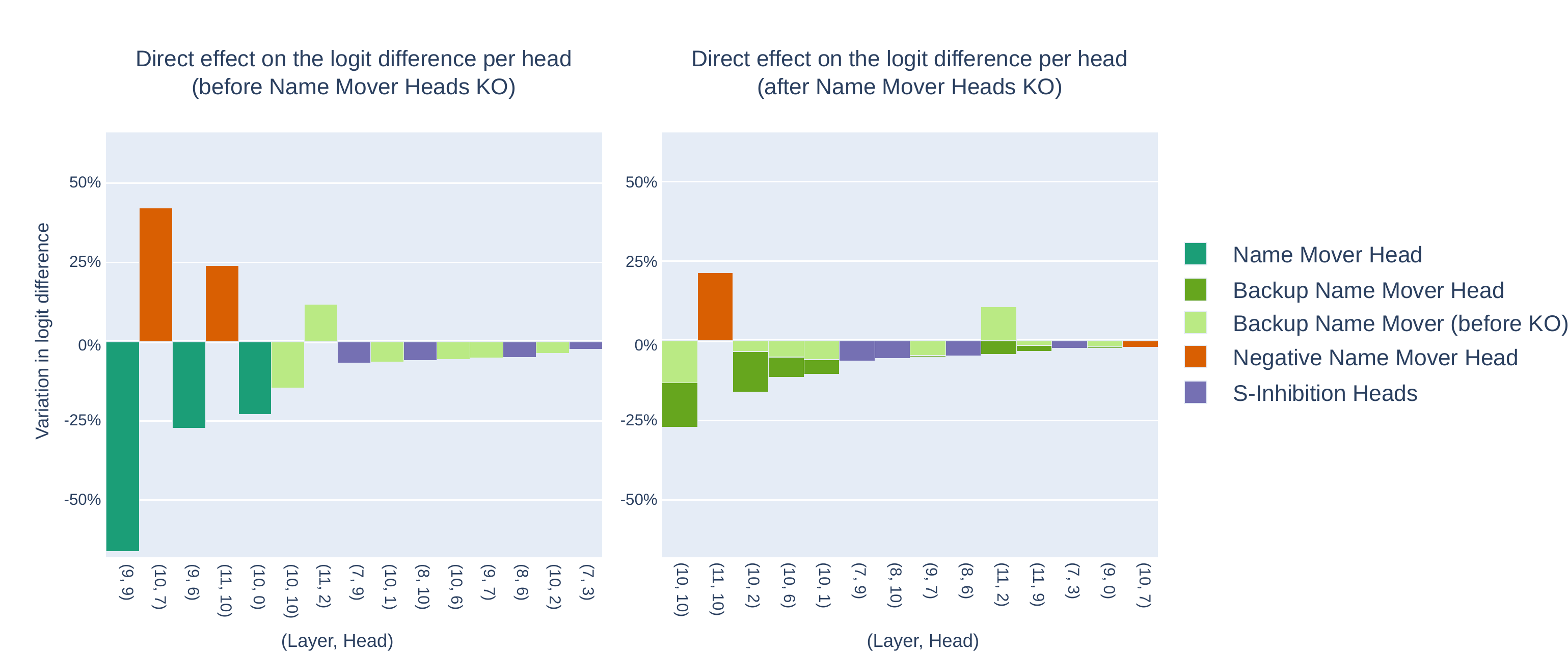}
     \caption{Discovery of the Backup Name Mover Heads. Left: results from the path patching experiment in Figure \ref{fig:name_moversB}. Right: the same path patching experiment results except computed after knocking out the Name Mover Heads. In both plots, the heads are ordered by decreasing order of the absolute value of their effect.}
     \label{fig:backup_nm_discovery}
\end{figure}

Here we discuss in more detail the discovery of the Backup Name Mover Heads. As shown in Figure \ref{fig:backup_nm_discovery}, knocking-out the three main Name Mover Heads surprisingly changes the behavior of the other heads that write in the IO-S direction (both positively and negatively). These heads compensate for the loss of function from the Name Mover Heads such that the logit difference is only 5\% lower. We observe that the Negative Name Mover Heads have a less negative effect on logit difference, and 10.7 even has a positive effect on the logit difference after the knockout. The other heads that affected slightly positively the logit difference before the knock-out become the main contributors. Both the reason and the mechanism of this compensation effect are still unclear. We think that this could be an interesting phenomenon to investigate in future works.

Among the heads influencing positively the logit difference after knockout, we identified S-Inhibition Heads and a set of other heads that we called \emph{Backup Name Mover Heads}. We arbitrarily chose to keep the eight heads that were not part of any other groups, and affected the logit difference with an effect size above the threshold of $2\%$. 

\begin{figure}
    \centering
    \includegraphics[width=\textwidth]{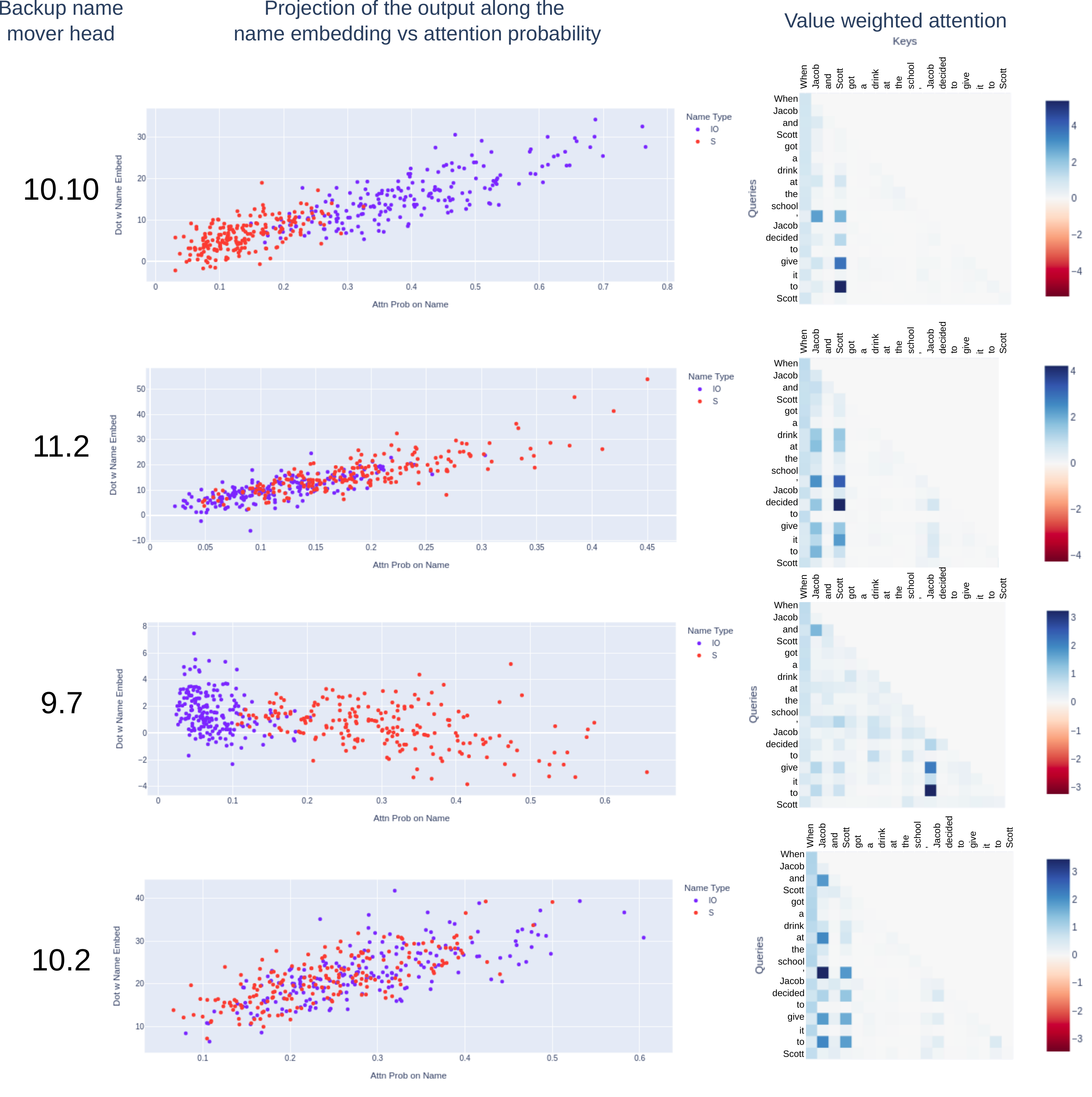}
    \caption{Four examples of Backup Name Mover Heads. Left: attention probability vs projection of the head output along $W_U[IO]$ or $W_U[S]$ respectively. Right: Attention pattern on a sample sequence.}
    \label{fig:backup_nm_analyisis}
\end{figure}

In Figure \ref{fig:backup_nm_analyisis} we analyze the behavior of those newly identified heads with similar techniques as Name Mover Heads. Those can be grouped in 4 categories.


\begin{itemize}
    \item Four heads (9.0, 10.1, 10.10 and 10.6) that behave similarly to Name Mover Heads in their attention patterns, and the scatter plots of attention vs. dot product of their output with $W_U[IO] - W_U[S]$ (e.g 10.10 in Figure \ref{fig:backup_nm_analyisis}).
    \item Two heads (10.2, 11.9) that pay equal attention to S1 and IO and write both of them (e.g 10.2 in Figure \ref{fig:backup_nm_analyisis}).
    \item One head, 11.2, that pays more attention to S1 and writes preferentially in the direction of $W_U[S]$.
    \item One head, 9.7, pays attention to S2 and writes negatively.
\end{itemize}

We did not thoroughly investigate this diversity of behavior, more work can be done to precisely describe these heads. However, these heads are also the ones with the less individual importance for the task (as shown by their minimality score in Figure \ref{fig:naive_minimality}). The exact choice of Backup Name Mover Heads doesn't change significantly the behavior of the circuit.


\section{GPT-2 small full architecture} \label{app:notation}

Here we define all components of the GPT-2 architecture, including those we don't use in the main text. GPT-2 small has the following parameters

\begin{compactitem}
\item $N$: number of input tokens.
\item $V$: length of vocabulary of tokens.
\item $d$: residual stream dimension.
\item $L$: number of layers.
\item $H$: number of heads per layer.
\item $D$: hidden dimension of MLPs.
\end{compactitem}

It uses layer norm, the non-linear function

\begin{equation}
\text{LN}(x) \eqdef \frac{x-\bar{x}}{\sqrt{\sum_i (x_i - \bar{x}_i )^2}},
\label{eqn:layer_norm_definition}
\end{equation}

where the mean and the difference from the mean sum are over all components of the dimension $d$ vector in each sequence position. This is then followed by a learned linear transformation $M$ (different for each layer norm).

In GPT-2 the MLPs all have one hidden layer of dimension $D$ and use the $\text{GeLU}$ non-linearity. Their input is the layer normed state of the residual stream.

We addressed the parametrisation of each attention head in the main text, and cover the technical details of the $W_{QK}$ and $W_{OV}$ matrix here: the attention pattern is $A_{i, j} = \text{softmax}( x^T W_{QK}^{i, j} x )$ where the softmax is taken for each token position, and is unidirectional. We then have $h_{i, j}(x) \eqdef M \circ \text{LN} ( (A_{i, j} \otimes W_{OV}^{i, j}) . x)$. 

Algorithm \ref{gpt2_algo} describe how these elements are combined in the forward pass of GPT-2 small.

\begin{algorithm}
\caption{GPT-2.}\label{gpt2_algo}
\begin{algorithmic}[1]
\Require $\text{Input tokens $T$; returns logits for next token.}$
\State $w \gets \text{One-hot embedding of T}$
\State $x_0 \gets W_E w \text{ (sum of token and position embeddings)} $
\For{$i=0$ to $L$}
    \State $y_i \gets 0 \in \mathbb{R}^{N \times d}$
    \For{$j=0$ to $H$}
        \State $y_i \gets y_i + h_{i, j}(x_i), \text{ the contribution of attention head $(i, j)$}$
    \EndFor
    \State $y_i' \gets m_i(y_i), \text{ the contribution of MLP at layer $i$}$ 
    \State $x_{i+1} \gets x_i + y_i + y_i' \text{ (update the residual stream)}$ 
\EndFor \\
\Return $W_U \circ M \circ \text{LN} \circ x_{L}$
\end{algorithmic}
\end{algorithm}

\section{Validation of the induction mechanism on sequences of random tokens} \label{app:random_tok_seq}

We run GPT-2 small on sequences of 100 tokens sampled uniformly at random from GPT-2's token vocabulary. Each sequence \texttt{A} was duplicated to form \texttt{AA}, a sequence twice as long where the first and second half are identical. On this dataset, we computed two scores from the attention patterns of the attention heads:
\begin{itemize}
    \item The \emph{previous token score}: we averaged the attention probability on the off-diagonal. This is the average attention from the token at position $i$ to position $i-1$.
    \item The \emph{induction score}: the average attention probability from $T_i$ to the token that comes after the first occurrence of $T_i$ (i.e. $T_{i-99}$)
\end{itemize} 

\begin{figure}
    \centering
    \includegraphics[width=\textwidth]{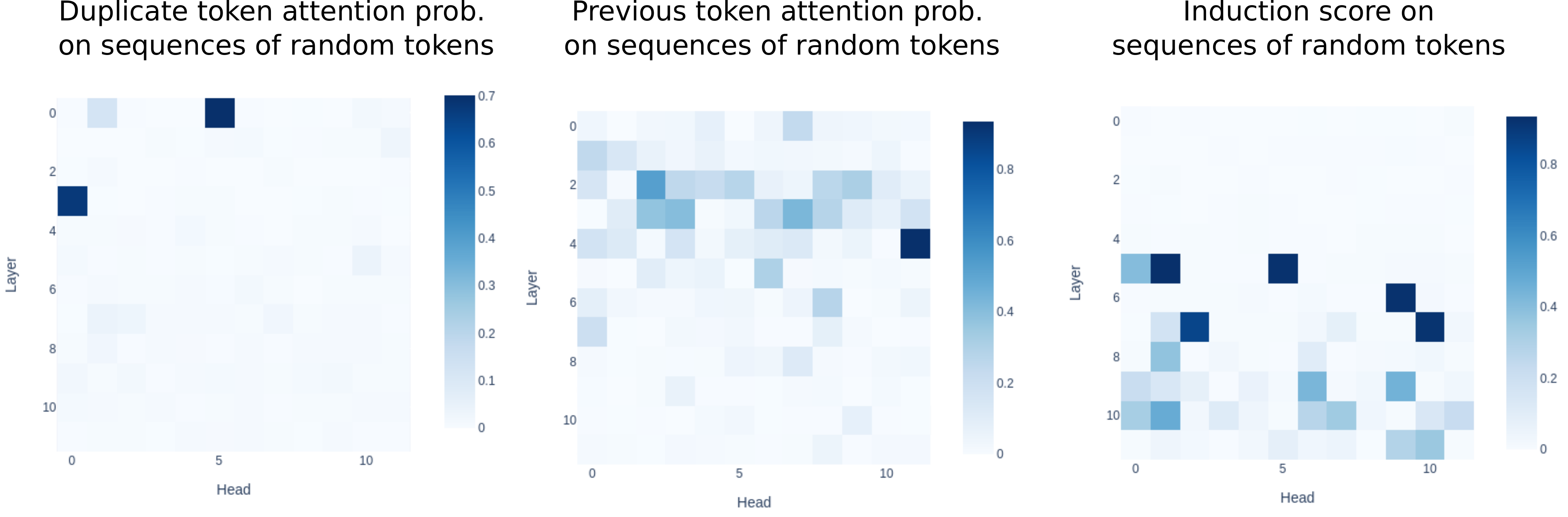}
    \caption{Attention scores on sequences of repeated random tokens. Left: Duplicate score, the average attention probability from a token to its previous occurrence. Center: Previous token attention score, it is the average of the off-diagonal attention probability. Right: Induction score. Average attention probability from the second occurrence of \texttt{[A]} to \texttt{[B]} on patterns \texttt{[A][B]...[A]}.}
    \label{fig:random_token}
\end{figure}

These two scores are depicted in Figure \ref{fig:random_token} (center and right) for all attention heads. 

\textbf{Previous Token Heads.} 4.11 and 2.2 are the two heads with the highest previous token score on sequences of random tokens. This is a strong validation of their role outside $\pioi$.

\textbf{Induction Heads.} \citet{olsson2022context} define an Induction Head according to its behavior on repeated sequences of random tokens. The attention head must demonstrate two properties. i) Prefix-matching property. The head  attends to \texttt{[B]} from the last \texttt{[A]} on pattern like \texttt{[A]\,[B]\,...\,[A]} ii) Copy property. The head contribute positively to the logit of \texttt{[B]} on the pattern \texttt{[A][B]...[A]}.

5.5 and 6.9 are among the 5 heads with the highest induction score. This validates their prefix-matching property introduced in \citet{olsson2022context}. 

\begin{figure}
    \centering
    \includegraphics[width=0.9\textwidth]{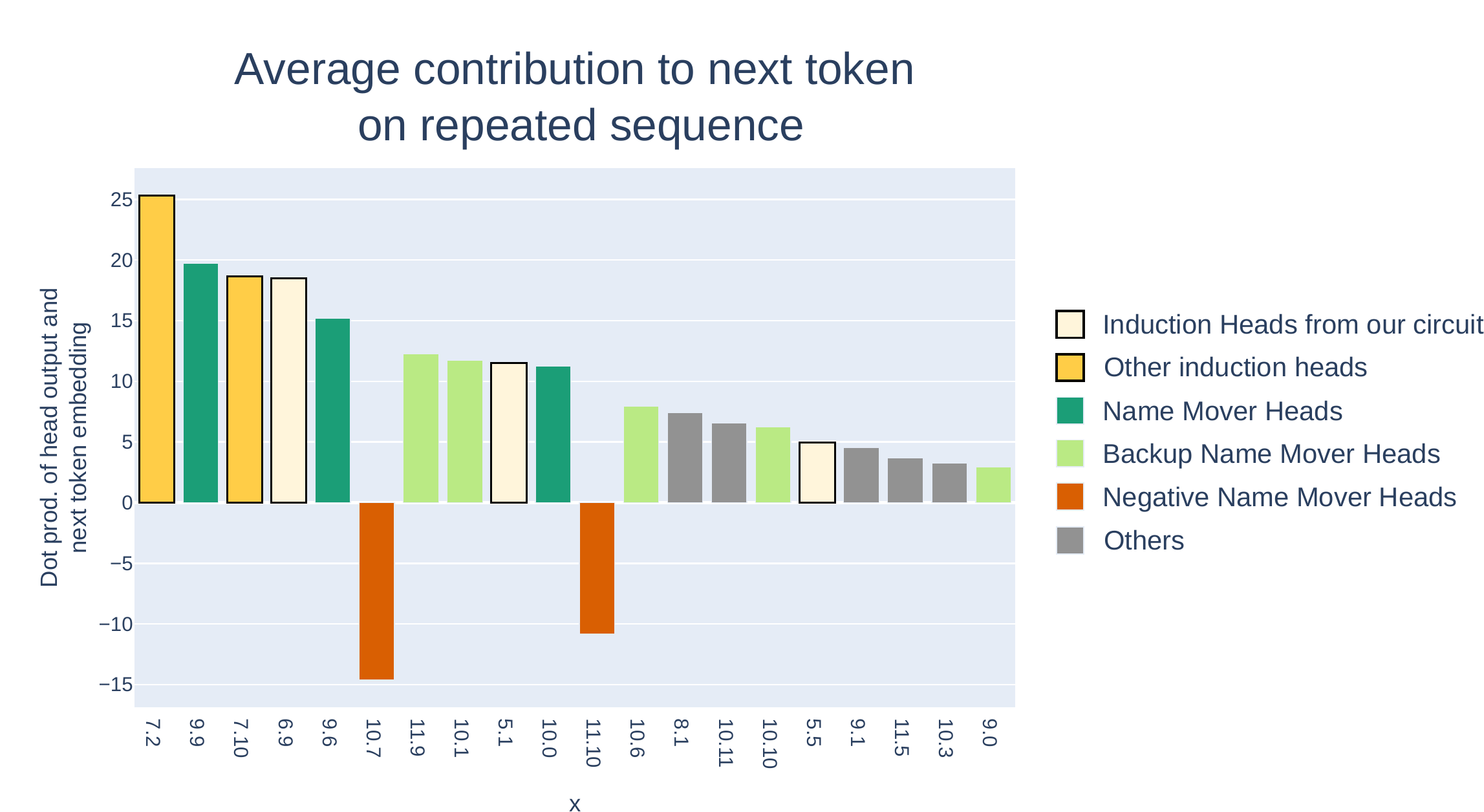}
    \caption{Contribution to the next token prediction per head on repeated sequences of tokens. The heads are ordered by decreasing absolute values of contribution. Black contour: heads with attention patterns demonstrating prefix matching property. }
    \label{fig:copy_property_ind}
\end{figure}

To check their copy property, we computed the dot product $\langle h_i(X), W_U[B] \rangle$ between the output of the head $h_i$ on sequence $X$ and the embedding of the token \texttt{[B]} on repeated sequences of random tokens. The results are shown in Figure \ref{fig:copy_property_ind}. The two Induction Heads (5.5 and 6.9) appear in the 20 heads contributing the most to the next token prediction. Thus validating their copying property. 

We also noticed that the majority of the Negative, Backup and regular Name Mover Heads appear to write in the next token direction on repeated sequences of random tokens, and Negative Name Movers Heads contribute negatively. This suggests that these heads are involved beyond the IOI task to produce next-token prediction relying on contextual information.

\section{Validation of Duplicate Token Heads} \label{app:validation_dups}

On the repeated sequences of random tokens (Appendix \ref{app:random_tok_seq}) we also computed the \emph{duplicate score}. For each token $T_i$ in the second half of a sequence, we average the attention probability from $T_i$ to its previous occurrence in the first half of the sequence (i.e. $T_{i-100}$). 

The duplicate token scores for all attention heads are depicted in Figure \ref{fig:random_token}. 3.0 and 0.1 are among the three heads with the highest duplicate token score (Figure \ref{fig:random_token}). This is evidence of their role of Duplicate Token Heads outside the circuit for the IOI task. 

However, the fuzzy Duplicate Head 0.10 doesn't appear on this test. By qualitatively investigating its attention patterns on Open Web Text, we found that this head attends strongly to the current token. Moreover, when the current token is a name, and it is duplicated, the head attends to its previous occurrence. 








\section{Role of MLPs in the task} 
\label{app:mlp}
\begin{figure}[H]
    \centering
    \includegraphics[width=0.9\textwidth]{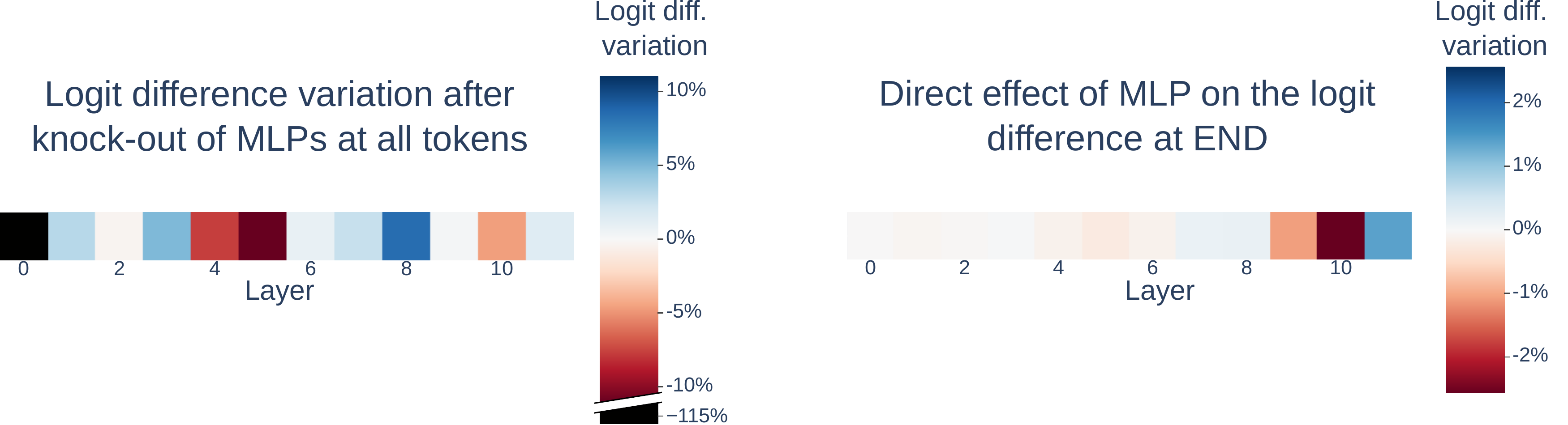}
    \caption{Left: change in logit difference from knocking out each MLP layer. Right: change in logit difference after a path patching experiment investigating the direct effect of MLPs on logits.}
    \label{fig:mlp_KO}
\end{figure}

In the main text, all of the circuit components are attention heads. Attention heads are the only modules in transformers that move information across token positions -- a crucial component of the IOI task -- so they were our main subject of interest. However, MLPs can still play a significant role in transforming the information in each residual stream. We explored this possibility by measuring the direct and indirect effects of each of the MLPs in Figure \ref{fig:mlp_KO}. In these experiments, for each MLP in turn, we did a path patching experiment (Section \ref{sec:s_inhibition}) to measure the direct effect and a knock-out experiment for the indirect effect.

We observe that MLP0 has a significant influence on logit difference after knock-out (it reverses the sign of the logit difference) but the other layers don't seem to play a big role when individually knocked out. When all MLP layers other than the first layer are knocked out, however, the logit difference becomes $-1.1$ (a similar effect to the knockout of MLP0 alone). 

\section{Minimality sets} \label{app:minimality}

The sets that were found for the minimality tests are listed in Figure \ref{fig:table_minimality}.

\begin{figure}
\centering
\begin{tabular}{ |c|c|c|c|c| } 
\hline
\textbf{$v$} & Class & \textbf{$K$} $\cup \{v\}$ & $F( C \setminus (K \cup \{v\}) )$ & $F( C \setminus K )$ \\ 
\hline \hline

(9, 9) & Name Mover & [(9, 9)] & 2.26 & 2.62 \\
\hline
(10, 0) & Name Mover & [(9, 9), (10, 0)] & 1.91 & 2.26 \\
\hline
(9, 6) & Name Mover & [(9, 9), (10, 0), (9, 6)] & 2.11 & 1.91 \\
\hline
(10, 7) & Negative & [(11, 10), (10, 7)] & 4.07 & 3.13 \\
\hline
(11, 10) & Negative & [(11, 10), (10, 7)] & 4.07 & 3.27 \\
\hline
(8, 10) & S Inhibition & [(7, 9), (8, 10), (8, 6), (7, 3)] & 0.24 & 1.01 \\
\hline
(7, 9) & S Inhibition & [(7, 9), (8, 10), (8, 6), (7, 3)] & 0.24 & 0.92 \\
\hline
(8, 6) & S Inhibition & [(7, 9), (8, 10), (8, 6), (7, 3)] & 0.24 & 0.86 \\
\hline
(7, 3) & S Inhibition & [(7, 9), (8, 10), (8, 6), (7, 3)] & 0.24 & 0.43 \\
\hline
(5, 5) & Induction & [(5, 9), (5, 5), (6, 9), (5, 8)] & 0.97 & 2.12 \\
\hline
(5, 9) & Induction & [(11, 10), (10, 7), (5, 9)] & 3.33 & 4.07 \\
\hline
(6, 9) & Induction & [(5, 9), (5, 5), (6, 9), (5, 8)] & 0.97 & 1.46 \\
\hline
(5, 8) & Induction & [(11, 10), (10, 7), (5, 8)] & 3.83 & 4.07 \\
\hline
(0, 1) & Duplicate Token & [(0, 1), (0, 10), (3, 0)] & 0.60 & 1.90 \\
\hline
(0, 10) & Duplicate Token & [(0, 1), (0, 10), (3, 0)] & 0.60 & 1.66 \\
\hline
(3, 0) & Duplicate Token & [(0, 1), (0, 10), (3, 0)] & 0.60 & 1.05 \\
\hline
(4, 11) & Previous Token & [(4, 11), (2, 2)] & 1.31 & 2.28 \\
\hline
(2, 2) & Previous Token & [(4, 11), (2, 2)] & 1.31 & 1.75 \\
\hline
(11, 2) & Backup Name Mover & All previous NMs and backup NMs & 0.95 & 1.37 \\
\hline
(10, 6) & Backup Name Mover & All previous NMs and backup NMs & 1.65 & 1.88 \\
\hline
(10, 10) & Backup Name Mover & All previous NMs and backup NMs & 1.88 & 2.11 \\
\hline
(10, 2) & Backup Name Mover & All previous NMs and backup NMs & 1.49 & 1.65 \\
\hline
(9, 7) & Backup Name Mover & All previous NMs and backup NMs & 0.81 & 0.95 \\
\hline
(10, 1) & Backup Name Mover & All previous NMs and backup NMs & 1.37 & 1.49 \\
\hline
(11, 9) & Backup Name Mover & All name movers and negative heads & 0.41 & 0.45 \\
\hline
(9, 0) & Backup Name Mover & All name movers and negative heads & 0.41 & 0.45 \\
\hline

\end{tabular}
\caption{$K$ sets for minimality for each $v$.}
\label{fig:table_minimality}
\end{figure}
\section{Template for adversarial examples} \label{app:advexes_templates}
The design of adversarial examples relies on adding a duplicate IO to the sentences. To this end, we used a modification of the templates described in appendix \ref{app:templates}. We added an occurrence of [A] in the form of a natural sentence, independent of the context. The list of sentence is visible in Figure \ref{fig:template_advexes}.

\begin{figure}[H]
\centering
\begin{tabular}{ |c| } 
\hline
[A] had a good day.\\
\hline
[A] was enjoying the situation.\\
\hline
[A] was tired.\\
\hline
[A] enjoyed being with a friend.\\
\hline
[A] was an enthusiast person.\\
\hline
\end{tabular}
\caption{Templates for the natural sentences used in the generation of adversarial examples. The sentences were chosen to be independent of the context.}

\label{fig:template_advexes}
\end{figure}
\section{Greedy Algorithm} \label{app:greedy}

The Algorithm \ref{greedy_algo} describes the procedure used to sample sets for checking the completeness criteria using greedy optimization. In practice, because the naïve and the full circuit are not of the same size, we chose respectively $k=5$ and $k=10$ to ensure a similar amount of stochasticity in the process. We run the procedure 10 times and kept the 5 sets with the maximal important incompleteness score (including the intermediate $K$).

\begin{algorithm}
\caption{The greedy sampling procedure for sets to validate the completeness citeria.}\label{greedy_algo}
\begin{algorithmic}[1]
\State $K \gets \emptyset$
\For{$i$ to $N$}
    \State Sample a random subset $V \subseteq C\setminus K$ of $k$ nodes uniformly.
    \State $v_{\text{MAX}}$ $\gets$ $\argmax_{v \in V} |F(C \setminus (K \cup \{v\})) - F( C\setminus K)|$
    \State $K \gets K \cup \{v_{\text{MAX}}\}$
\EndFor \\
\Return $K$
\end{algorithmic}
\end{algorithm}

As visible in Figure \ref{fig:table_complteness} the sets found by the greedy search contain a combination of nodes from different classes. This means that it is difficult to interpret how our circuit is incomplete, as mentioned in the main text. 

\begin{figure}
\centering
\begin{tabular}{ |c| } 
\hline
$K$ found by greedy optimization \\ 
\hline \hline
(9, 9), (9, 6), (5, 8), (5, 5), (2, 2) \\
\hline
(9, 9), (11, 10), (10, 7), (8, 6), (5, 8), (4, 11) \\
\hline
(10, 7), (5, 5), (2, 2), (4, 11)\\
\hline
(9, 9), (11, 10), (10, 7), (11, 2), (3, 0), (5, 8), (2, 2) \\
\hline

\end{tabular}
\caption{4 sets $K$ found by the greedy optimization procedure on our circuit.}
\label{fig:table_complteness}
\end{figure}

\end{document}